\def\cvprPaperID{6762}
\def\confName{ICCV}
\def\confYear{2023}

\newif\ifreview 
\newif\ifarxiv \newcommand{\arxiv}{\arxivtrue}
\newif\ifcamera 
\newif\ifrebuttal 

\arxiv

\pdfoutput=1
\documentclass[10pt,twocolumn,letterpaper]{article}
\ifreview \usepackage[review]{cvpr} \fi
\ifarxiv \usepackage[pagenumbers]{cvpr} \fi
\ifrebuttal \usepackage[rebuttal]{cvpr} \fi
\ifcamera \usepackage{cvpr} \fi

\usepackage{graphicx}
\usepackage{amsmath}
\usepackage{amssymb}
\usepackage{booktabs}

\usepackage{times}
\usepackage{microtype}
\usepackage{epsfig}
\usepackage[table,xcdraw]{xcolor}
\usepackage{caption}
\usepackage{float}
\usepackage{placeins}
\usepackage{color, colortbl}
\usepackage{stfloats}
\usepackage{enumitem}
\usepackage{tabularx}
\usepackage{xstring}
\usepackage{cuted} 
\usepackage{multirow}
\usepackage{xspace}
\usepackage{url}
\usepackage{subcaption}
\usepackage{xcolor}
\usepackage[hang,flushmargin]{footmisc}

\ifcamera \usepackage[accsupp]{axessibility} \fi

% ...

\ifarxiv  \fi

\newcommand{\R}[1]{{%
    \textbf{%
        \ifstrequal{#1}{1}{\textcolor{red}{R#1}}{%
        \ifstrequal{#1}{2}{\textcolor{blue}{R#1}}{%
        \ifstrequal{#1}{3}{\textcolor{magenta}{R#1}}{%
        \ifstrequal{#1}{4}{\textcolor{teal}{R#1}}{%
                           \textcolor{cyan}{R#1}%
        }}}}%
    }%
}}

\definecolor{SigmaColor}{rgb}{0.98,0.45,0.0}

\usepackage{xr-hyper}

\makeatletter
\newcommand*{\addFileDependency}[1]{
  \typeout{(#1)}
  \@addtofilelist{#1}
  \IfFileExists{#1}{}{\typeout{No file #1.}}
}

\makeatother

\usepackage[pagebackref,breaklinks,colorlinks]{hyperref}
\usepackage[capitalize]{cleveref}
\crefname{section}{Sec.}{Secs.}
\crefname{table}{Table}{Tables}
\crefname{figure}{Fig.}{Figs.}

\begin{document}

\title{NCHO: Unsupervised Learning for \\Neural 3D Composition of Humans and Objects}
\author{Taeksoo Kim$^{1}$ \; Shunsuke Saito$^2$ \; Hanbyul Joo$^{1}$\\
$^1$Seoul National University \quad $^2$Meta Reality Labs \\
{\tt\small taeksu98@snu.ac.kr \quad shunsuke.saito16@gmail.com \quad hbjoo@snu.ac.kr \quad} \\
{\tt\small \href{https://taeksuu.github.io/ncho}{https://taeksuu.github.io/ncho}}
}
\maketitle
\begin{strip}\centering
\vspace{-1.8cm}
\includegraphics[width=\linewidth, trim={0 0 0 0.2cm},clip]{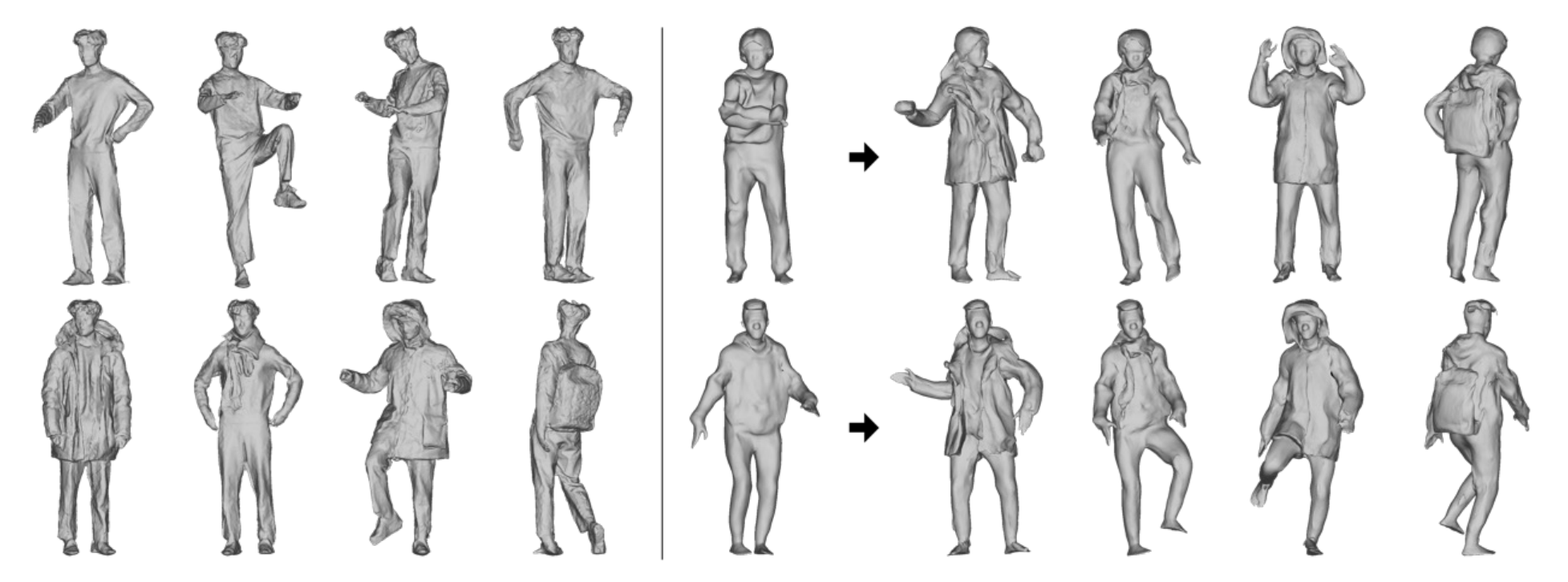}
\vspace{-9mm}
\captionof{figure}{From 3D scans of the ``source human'' in casual clothing (left top) and with additional outwear or objects (left bottom), our method automatically decomposes objects from the source human and builds a compositional generative model that enables 3D avatar creations of novel human identities with variety of outwear and objects (right) in an unsupervised manner. }
\label{fig:teaser}
\end{strip}

\begin{abstract}
Deep generative models have been recently extended to synthesizing 3D digital humans. However, previous approaches treat clothed humans as a single chunk of geometry without considering the compositionality of clothing and accessories. As a result, individual items cannot be naturally composed into novel identities, leading to limited expressiveness and controllability of generative 3D avatars. 
While several methods attempt to address this by leveraging synthetic data, the interaction between humans and objects is not authentic due to the domain gap, and manual asset creation is difficult to scale for a wide variety of objects. 
In this work, we present a novel framework for learning a compositional generative model of humans and objects (backpacks, coats, scarves, and more) from real-world 3D scans. Our compositional model is interaction-aware, meaning the spatial relationship between humans and objects, and the mutual shape change by physical contact is fully incorporated. The key challenge is that, since humans and objects are in contact, their 3D scans are merged into a single piece. To decompose them without manual annotations, we propose to leverage two sets of 3D scans of a single person with and without objects. Our approach learns to decompose objects and naturally compose them back into a generative human model in an unsupervised manner. 
Despite our simple setup requiring only the capture of a single subject with objects, our experiments demonstrate the strong generalization of our model by enabling the natural composition of objects to diverse identities in various poses and the composition of multiple objects, which is unseen in training data.
\end{abstract}

\vspace{-6mm}
\section{Introduction}
\label{sec:intro}
Generative modeling of 3D humans from real-world data has shown promise to represent and synthesize diverse human shapes, poses, and motions. Especially, the ability to create realistic humans in diverse clothing and accessories (e.g. backpacks, scarves, and hats) is indispensable for a myriad of applications including VR/AR, entertainment, and virtual try-on. The early work~\cite{anguelov2005scape, pavlakos19expressive, SMPL:2015, joo2018total, xu2020ghum} has demonstrated success in modeling undressed human bodies from real-world scans. Recently, the research community has been focused on the generative modeling of clothed humans~\cite{ma2021pop, chen2022gdna, corona2021smplicit}, to better represent humans in everyday life.

Recent advancements in shape representations such as Neural Fields~\cite{xie2022neural} mitigate the need for pre-defining topology or template of clothing, enabling to build animatable clothed humans from raw 3D scans~\cite{saito2021scanimate, chen2021snarf}. Along with its advantage in strong expressive power for avatar modeling, this approach also allows the models to learn faithful interactions between objects and humans. However, since raw 3D scans do not provide a clear separation of different components, existing approaches typically treat humans, clothing, and accessories as an entangled block of geometry~\cite{chen2022gdna}. In this paper, we argue that this leads to suboptimal expressiveness and composability of the generative avatars. Many applications require more intuitive control to add, replace, or modify objects while maintaining human identity. 
To make avatars explicitly compositable with objects, some approaches propose to leverage synthetic data~\cite{bhatnagar2019multi, jiang2020bcnet, corona2021smplicit}. However, the manual creation of 3D assets remains a challenge and is extremely difficult to scale. Moreover, the physical interaction of bodies, clothing, and accessories in synthetic data tends to be less faithful due to the domain gap.

In contrast to prior methods, our goal is to build a compositional generative model of objects and humans from real-world observations. The core challenge lies in the difficulty of learning the composition and decomposition of objects in contact from raw 3D scans. Capturing objects in isolation does not lead to faithful composition due to the lack of realistic deformations induced by physical contact. Thus, while it is essential to collect 3D scan data on objects and humans in contact, the joint scanning of humans with objects only provides an entangled block of 3D geometry as mentioned, and accurately segmenting different components requires non-trivial 3D annotation efforts. 

Upon these challenges, our contributions are: scalable data capture protocol, unsupervised decomposition of objects and humans, and generalizable neural object composition.

\noindent\textbf{Scalable Data Capture}. Capturing multiple identities with various poses and objects requires prohibitively large time and storage. To overcome this issue, we propose to collect human-object interactions with diverse poses only from a single subject, referred to as the ``source human''. To enable the decomposition of objects, we also capture the same person without any objects, where the deviation between two sets defines ``objects'' in our setup. Examples are shown in Fig.~\ref{fig:dataset}. This capture protocol offers sufficient diversity in poses and object types within a reasonable capture time.

\noindent\textbf{Unsupervised Decomposition of Objects}. To separate objects from the source human, we leverage the expressiveness of the generative human model based on implicit surface representation~\cite{chen2022gdna}. We train a human module without objects, and then jointly optimize the latent codes of the avatar and a generative model for objects to best explain the 3D scans of the person with objects. While the human module accounts for state differences in pose and clothing, the object-only module learns to synthesize the residual geometry as an object layer in an unsupervised manner. Notably, objects in our work are defined as residual geometry that cannot be explained by the trained human-only module.

\noindent\textbf{Neural Object Composition}. While the unsupervised decomposition successfully separates objects from the source human, we observe that naively composing it to novel identities from other datasets~\cite{renderpeople, tao2021function4d} leads to undesired artifacts and misalignment in the contact regions. To address this, we propose a neural composition method by introducing another composition MLP that takes latent features from both human and object modules to make a final shape prediction. Due to the local nature of MLPs, our approach plausibly composes objects to novel identities without retraining as in Fig.~\ref{fig:teaser}.

Our experiments show that our compositional generative model is superior to existing approaches without explicit disentanglement of objects and humans~\cite{chen2022gdna}. In addition, we show that our model can be used for fine-grained controls including object removal from 3D scans and multiple object compositions on a human, demonstrating the utility and expressiveness of our approach beyond our training data.

\section{Related Work}
\label{sec:related}
\noindent \textbf{3D Human Models.}
Representing plausible 3D human bodies while handling diverse variations in shapes and poses is a long-standing problem. Due to the challenge in modeling diverse shape variation, the early work~\cite{anguelov2005scape, joo2018total, SMPL:2015, pavlakos19expressive, xu2020ghum} mainly focuses on the undressed 3D human body by learning mesh-based statistical models deformed from a template mesh.
To model dressed 3D humans, the follow-up work~\cite{alldieck2018video, alldieck2019learning, ma2020learning} adds 3D offsets on top of the parametric undressed human body models to represent clothing. Yet, the topological constraints and the resolution of the template model restrict these methods from modeling arbitrary shapes of clothing with high-frequency details.
Recently emerging deep implicit shape representation~\cite{chen2019learning, mescheder2019occupancy, park2019deepsdf, mildenhall2020nerf} provides a breakthrough in expressing 3D humans by leveraging neural networks for representing continuous 3D shape space, where its efficacy is demonstrated in reconstructing clothed humans with high-fidelity from images~\cite{saito2019pifu, saito2020pifuhd, xiu2022icon}. There also has been an actively growing field to represent animatable 3D human avatars using 3D scans~\cite{deng2020nasa, chen2021snarf, saito2021scanimate, tiwari2021neural, hong2022avatarclip, mihajlovic2022coap, mihajlovic2021leap, chen2022gdna}. However, prior 3D human models have paid little attention to the joint modeling of humans and objects in close contact.

\noindent \textbf{2D/3D Generative Models.}
Generative models intend to express the plausible variations over the latent space, which can be used to create diverse realistic samples. 
There have been extensive studies in 2D generative modeling to create realistic photos~\cite{karras2017progressive, karras2019style, karras2020analyzing} via generative adversarial networks (GANs)~\cite{nips2014gan, goodfellow2020generative}, variational autoencoders (VAEs)~\cite{kingma2013auto}, and more recently, diffusion models~\cite{Sohl-ICML-2015dpm, Dhariwal-NIPS-2021guideddiffusion, ho-2022cascaded,Rombach-CVPR-2022ldm}.
Generative 3D modeling has also been actively explored. By leveraging the availability of a large-scale 3D object scans~\cite{chang2015shapenet},
many approaches present generative models for 3D objects~\cite{chen2019learning,mescheder2019occupancy, nguyen2019hologan,niemeyer2021giraffe,park2019deepsdf,schwarz2020graf,chan2022efficient}.
Relatively few approaches have been presented for generative 3D human modeling, due to the lack of available 3D datasets for humans~\cite{xu2020ghum, alldieck2021imghum, corona2021smplicit, ma2020learning, chen2022gdna}. We show that our scalable data capture protocol and compositional generative model enable the synthesis of 3D humans with diverse objects in novel poses.

\noindent \textbf{Compositional Models.}
Compositional generative models via neural networks have been explored to represent different components as independent models, representing a whole scene by compositing them together. These approaches pursue controlling or sampling one component without affecting the rest. The early approaches focus on building such models in 2D for creating realistic 2D images via generative models~\cite{zhu2015learning, tsai2017deep, lin2018st, azadi2020compositional}. 
More recent approaches explore the compositional reasoning for 3D~\cite{niemeyer2021giraffe, yu2022unsupervised,wu2022object, yang2021learning, ost2021neural, br2022gcorf, wang2021learning, 
 li2023megane}. Most approaches in this direction aim at synthesizing realistic novel views by compositing NeRFs~\cite{mildenhall2020nerf} for 3D objects and scenes~\cite{niemeyer2021giraffe,wu2022object, yang2021learning} and for human faces~\cite{niemeyer2021giraffe, xue2022giraffe, br2022gcorf}. However, these approaches do not consider mutual shape deformations between objects. Human bodies are also treated as a composition of multiple body parts. These approaches attain final composition output by either max-pooling the outputs of individual components~\cite{deng2020nasa, mihajlovic2022coap} or by using another neural network~\cite{alldieck2021imghum, biswas2021hierarchical, su2021nerf, noguchi2021neural}.
While a recent work shows interaction-aware 3D composition reasoning is possible for faces and eyeglasses with extensive annotations and data preprocessing~\cite{li2023megane}, our approach supports diverse object categories without requiring any manual annotations.

\noindent \textbf{Garment Modeling.}
Due to the deformable nature of garments, capturing and modeling 3D clothing is challenging. Only a few 3D garment datasets have been presented~\cite{bhatnagar2019multi, zhu2020deep}, where laborious segmentation and post-processing are required to separate the garments from dummies or human bodies.
While most methods reconstruct a clothed 3D human as a single chunk of geometry~\cite{saito2019pifu, saito2020pifuhd, xiu2022icon}, there exist methods reconstructing the 3D clothing as a separate layer on top of parametric mesh model (e.g., SMPL) using segmentation~\cite{Feng2022scarf} or synthetic 3D assets~\cite{jiang2020bcnet, corona2021smplicit}. Virtual try-on has also been actively explored in graphics via physics simulation~\cite{xiang2022dressing} and or synthetic data~\cite{vidaurre2020fully}. In contrast, our approach learns a generative clothing and accessory model from real-world observations in an unsupervised fashion.

\section{Preliminaries}

\begin{figure}[t]
\includegraphics[trim={1cm 1.6cm 1cm 1.5cm},clip,width=1.0\columnwidth]{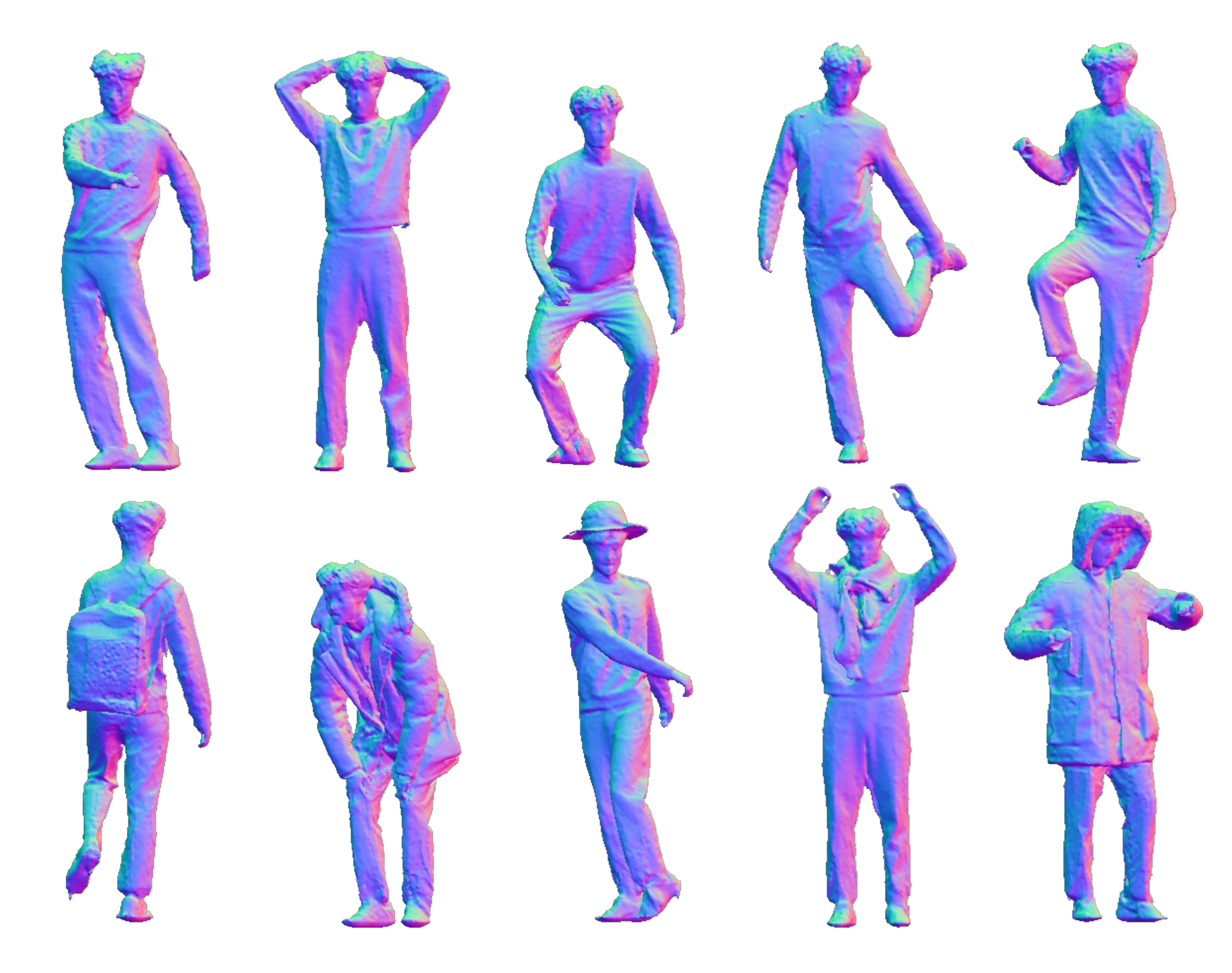}
\vspace{-20px}
\caption{\textbf{Examples of Our Datasets.} Top row: sample scans of $\mathcal{S}_{sh}$ containing the source human without objects. Bottom row: sample scans of $\mathcal{S}_{sh+o}$ containing the source human with objects.}
\label{fig:dataset}
\vspace{-15px}
\end{figure}

\paragraph{Data Acquisition.}
\label{subsection:data_acq}
To model humans and objects in contact, we capture two sets of datasets, $\mathbf{S}_{sh}$ and $\mathbf{S}_{sh+o}$. $\mathbf{S}_{sh}$ consists of 3D scans of a single identity, denoted as ``source human" with various poses.
$\mathbf{S}_{sh+o}$ consists of 3D scans of the source human with a variety of objects or additional outwear as shown in Fig.~\ref{fig:dataset}. 
In this work, we choose coats, vests, backpacks, scarves, and hats to demonstrate the generality of our approach for outwear and everyday accessories. 
To support the generative modeling of objects, we capture multiple objects in each category.
In addition to $\mathbf{S}_{sh}$ and $\mathbf{S}_{sh+o}$, we also use other 3D human dataset~\cite{tao2021function4d} to train another target generative human model for composition, denoted $\mathbf{S}_{th}$.

We collect 3D scans with a system with synchronized and calibrated 8 Azure Kinects (see supp. mat. for details). 
We apply KinectFusion~\cite{newcombe2011kinectfusion} to fuse the depth maps, and then reconstruct watertight meshes with screened-poisson surface reconstruction~\cite{kazhdan2013screened}.
We also detect 2D keypoints using OpenPose~\cite{cao2017realtime} and apply the multi-view extension of SMPLify~\cite{Bogo2016} to obtain SMPL parameters~\cite{SMPL:2015} for each scan.

\vspace{-14px}
\paragraph{Generative Articulated Models.}
We adopt the generative human model~\cite{chen2022gdna} which
extends forward skinning with root finding~\cite{chen2021snarf} for cross-identity modeling. We briefly discuss the framework and highlight our key modifications.
The key idea in gDNA~\cite{chen2022gdna} is to represent occupancy fields conditioned by identity-specific latent codes $\mathbf{z}$ in a canonical space, and transform them into a posed space using forward linear blend skinning (LBS). The occupancy field defined for the  location $\mathbf{x}^c$ of a person in the canonical space can be represented as follows:
\begin{gather}
    o(\mathbf{x}^c) = O(\mathbf{x}^c, G(\mathbf{z})),
\label{eq:gdna}
\end{gather}
where $G(\cdot)$ is a spatially varying feature generator taking the latent code. While the original work~\cite{chen2022gdna} uses 3D feature voxels for the output of $G$, we use a tri-plane feature representation~\cite{chan2022efficient}, which achieves better performance with higher memory efficiency. The generated feature map is conditioned on the latent code $\mathbf{z}$ via adaptive instance normalization~\cite{huang2017arbitrary}.

To query the occupancy fields in a posed space point $\mathbf{x}^d$, we transform the canonical coordinate $\mathbf{x}^c$ as follows:
\begin{align}
\label{eq:warp}
    \mathbf{x}^d =\sum_{i=1}^{n_b} W_i(N(\mathbf{x}^c, \beta), \mathbf{z}) \cdot \mathbf{B}_i(\beta, \theta) \cdot \mathbf{x}^c,
\end{align}
where $W_i$ is the identity conditioned skinning network, which outputs LBS skinning weights for the $i$-th bone, and $N$ is the warping network given SMPL shape parameters $\beta \in \mathbb{R}^{10}$. $\mathbf{B}_i(\beta, \theta)$ is the transformation of the $i$-th bone in SMPL model given SMPL pose parameters $\theta \in \mathbb{R}^{24 \times 3}$ and $\beta$. To jointly learn the occupancy and deformation networks, we solve for $\mathbf{x}^c$ in Eq.~\ref{eq:warp} given $\mathbf{x}^d$ using iterative root finding~\cite{chen2021snarf}. We discard the surface normal prediction networks used in~\cite{chen2022gdna} in both canonical space and screen space.
Instead of hallucinating details with fake normals, we propose to model detailed geometry by jointly representing shapes as SDF together with the occupancy fields. 
As we can directly supervise SDF on surface normals~\cite{gropp2020implicit}, we model detailed geometry as true surface. 
However, we empirically find that directly replacing the occupancy with SDF leads to unstable training.
To mitigate instability, we propose a hybrid modeling of occupancy and SDF.
We disable the backpropagation of gradients from SDF to the deformation networks so that it is only supervised by the occupancy head.
See supp. mat. for details.

\begin{figure*}[t]
\includegraphics[width=\linewidth, trim={0 0 0 0},clip]{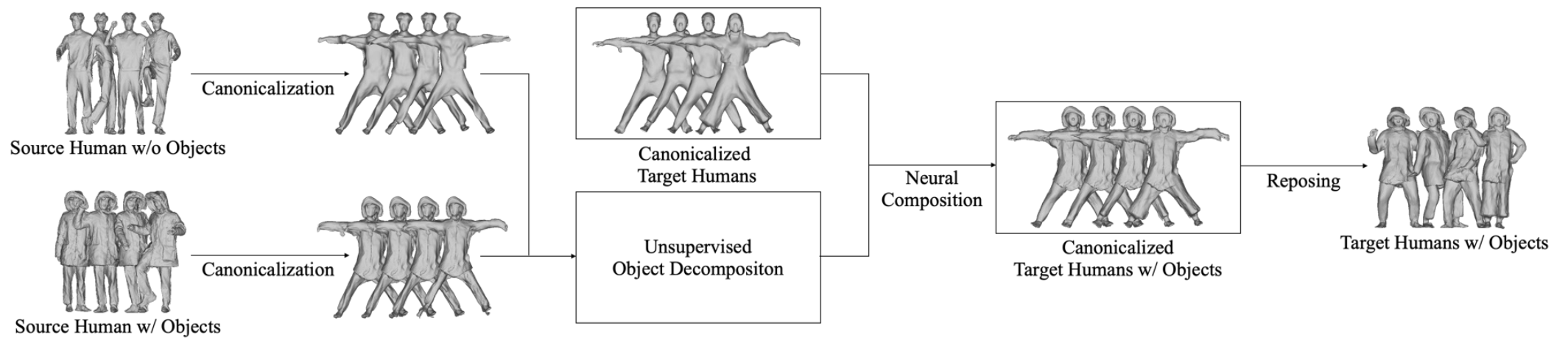}
\vspace{-18px}
\caption{\textbf{Overview.} From captured scans of the source human with and without objects, our method succesfully decomposes objects from humans without any supervision, allowing a generative model to learn the shapes of various objects. These objects are then added to novel identities via neural composition, resulting in the creation of diverse human avatars with controllable objects. }
\label{fig:overview}
\vspace{-18px}
\end{figure*}

\section{Method}
Our goal is to build a compositional generative model that composes generative objects on target humans from raw 3D scans. To this end, we introduce a generative human module and a generative object module, followed by a composition module. Fig.~\ref{fig:overview} shows an overview of our pipeline.

\vspace{-12px}
\paragraph{Human Module.} The human module $\mathcal{M}_{h} = (G_h, O_h, F_h, D_h)$ represents the geometry of the human part and it is composed of a feature generator $G_h$, a decoder $O_h$, and deformation networks $D_h= (W_h, N_h)$, where $W_h$ and $N_h$ are a skinning weight network and a warping network as in Eq.~\ref{eq:warp}.  As an output, $\mathcal{M}_{h}$ produces an occupancy value $o_h$, a feature vector $\mathbf{f}_h$, and a signed distance $d_h$ in the canonical space:
\begin{gather}
    (o_h, \mathbf{f}_{h}) = O_h(\mathbf{x}^c, G_h(\mathbf{z}_h)), \\
    d_h = F_h(\mathbf{x}^c, G_h(\mathbf{z}_h)).
\end{gather}
$\mathbf{f}_h$ is the intermediate latent feature before the last layer, and $\mathbf{z}_h$ is a learnable latent code to vary the geometry of the human part. 
Note that the hybrid modeling of occupancy and SDF is applied only to the human module as our losses for unsupervised object decomposition require occupancy. 

\vspace{-12px}
\paragraph{Object Module.} The object module $\mathcal{M}_{o} = (G_{o}, O_{o})$ is responsible for modeling the geometry of the object part. Since the object module and the human module share the same canonical space, the object module does not require separate deformation networks. $\mathcal{M}_{o}$ returns an occupancy value $o_o$, and a feature vector $\mathbf{f}_o$, which is the intermediate latent feature before the last layer, in the canonical space:
\begin{gather}
    (o_{o}, \mathbf{f}_{o}) = O_{o}(\mathbf{x}^c, G_{o}(\mathbf{z}_{o})),
\end{gather}
where $\mathbf{z}_o$ is a learnable latent code to vary the geometry of the object part.

\subsection{Neural Object Composition}

Since the outputs of the human module and the object module share the same canonical space and deformation networks, 
compositing the occupancy of the human and object modules in a closed-form~\cite{deng2020nasa, mihajlovic2022coap} is possible. However, we observe that this leads to misalignment in the contact regions and floating artifacts. To address these issues, we introduce a neural composition module parameterized by MLPs.

\begin{figure}[t]
\includegraphics[width=1\columnwidth]{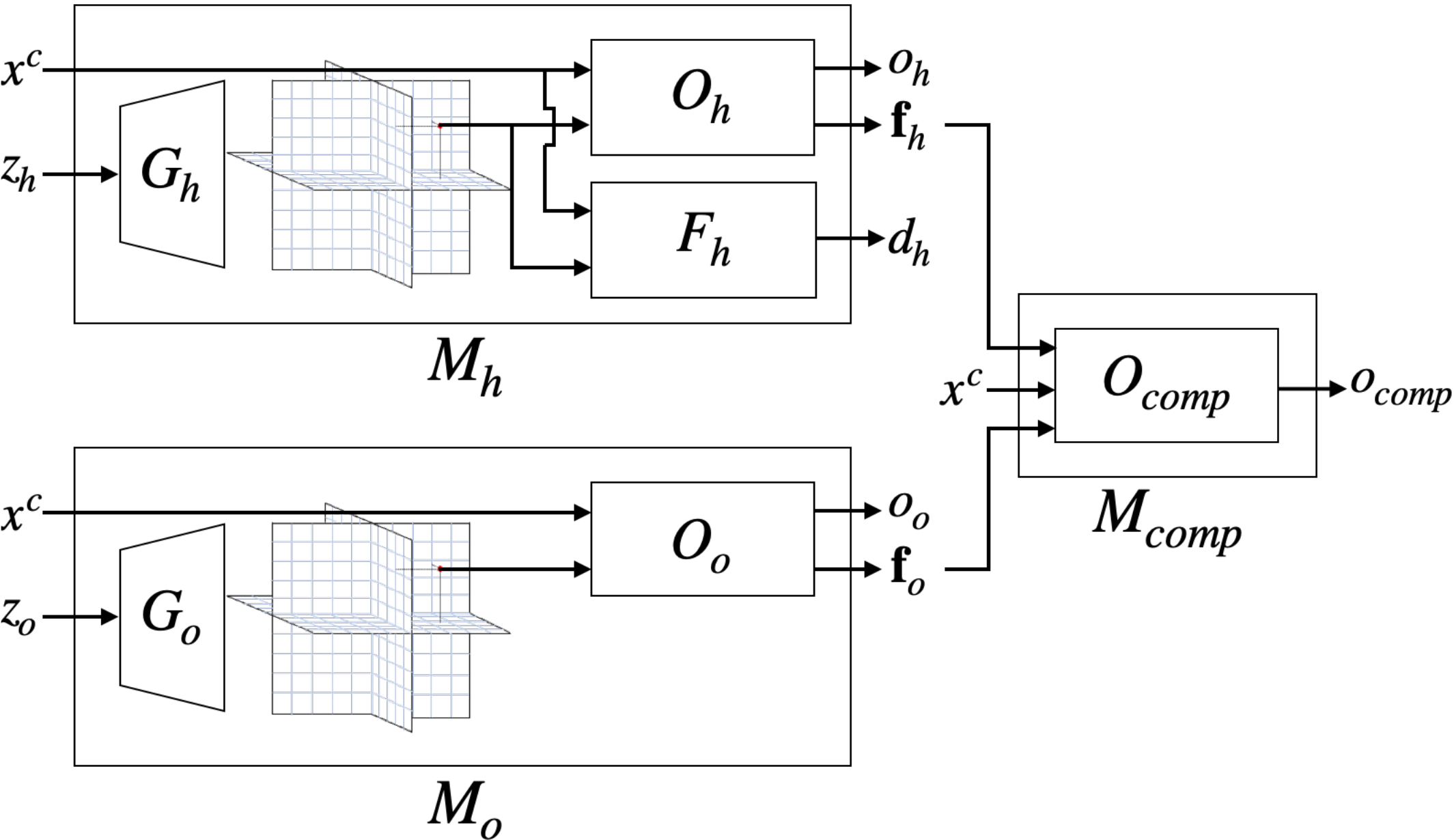}
\vspace{-15px}
\caption{\textbf{Model.} Given latent code $\mathbf{z}_h$, $\mathcal{M}_h$ predicts the occupancy fields and SDFs for humans in canonical space. Similarly, with latent code $\mathbf{z}_o$, $\mathcal{M}_o$ predicts the occupancy fields for objects. The features $\mathbf{f}_h$ and $\mathbf{f}_o$ from each network are passed to $\mathcal{M}_{comp}$ to predict the occupancy fields for final compositional outputs of humans and objects in the same canonical space.}
\label{fig:model}
\vspace{-18px}
\end{figure}

The composition module $\mathcal{M}_{comp} = (O_{comp}, D_{comp})$ is used to integrate humans and objects in the canonical space. We directly feed the feature vectors $\mathbf{f}_h$ and $\mathbf{f}_{o}$ from the human module and object module respectively as inputs. $\mathcal{M}_{comp}$ outputs the final occupancy value $o_{comp}$, after composition in the canonical space:
\begin{gather} 
\label{eq:comp}
o_{comp} = O_{comp}(\mathbf{x}^c, \mathbf{f}_h, \mathbf{f}_{o})
\end{gather}
Similar to the human module, the deformation networks $D_{comp}=(W_{comp}, N_{comp})$ provide the mapping from the canonical space to the posed space. The entire model is illustrated in Fig.~\ref{fig:model}.

\subsection{Unsupervised Object Decomposition} 
\label{sec:decomposition}

To decompose object layers from raw 3D scans in an unsupervised manner, our key idea is to represent objects as the residual of human geometry. To this end, we first train the human module $\mathcal{M}_{h}$ using $\mathbf{S}_{sh}$, the dataset of source human without objects, along with the learnable shape code $\mathbf{z}_{sh}$ for each scan. This allows the human module to account for slight shape variations of the source human by changing $\mathbf{z}_{sh}$.
In the next step, using $\mathbf{S}_{sh+o}$, the dataset of source human with objects, we jointly train all modules together. In particular, we freeze the human module $\mathcal{M}_{h}$ while optimizing $\mathbf{z}_{sh}$, $\mathcal{M}_{o}$, $\mathbf{z}_{o}$, and $\mathcal{M}_{comp}$. Intuitively, the pretrained human module tries to handle the geometry of the human part via optimization of $\mathbf{z}_{sh}$, while the object parts, which cannot be expressed by $\mathcal{M}_{h}$, are handled by $\mathcal{M}_{o}$ and $\mathbf{z}_{o}$.
Given the composed occupancy $o_{comp}$ in Eq.~\ref{eq:comp} and the predicted occupancy of the human module $o_{h}$, the target occupancy of the object module can be computed as $(1 - o_h) \cdot o_{comp}$. We jointly optimize the neural composition module and the object module $\mathcal{M}_{o}$ in an end-to-end manner using the loss functions discussed in Sec.~\ref{sec:training}.

\subsection{Training}
\label{sec:training}

Our system is trained using the datasets $\mathbf{S}_{th}$, $\mathbf{S}_{sh}$ and $\mathbf{S}_{sh+o}$ with their SMPL shape and pose parameters. Following the auto-decoding framework of~\cite{park2019deepsdf}, we jointly optimize the latent code $\mathbf{z}$ assigned for each scan along with the network weights during training. Every scan in each dataset is assigned its own latent code, denoted $\mathbf{z}_{th} \in \mathbb{R}^{L_{th}}$ for scans in $\mathbf{S}_{th}$, $\mathbf{z}_{sh} \in \mathbb{R}^{L_{sh}}$ for scans in $\mathbf{S}_{sh}$ and $\mathbf{z}_{o} \in \mathbb{R}^{L_{o}}$ for scans in $\mathbf{S}_{sh+o}$. For $\mathbf{z}_{o}$, we use one-hot encoding for each object category using the first 5 bits to enable random sampling from a specific category. Note that all latent codes are initialized with zero. 

To allow the unsupervised decomposition of objects from the source human as discussed in Sec.~\ref{sec:decomposition}, and to enable the creation of novel human identities with objects, we train two separate human modules $\mathcal{M}_{sh}$ and $\mathcal{M}_{th}$. $\mathcal{M}_{sh}$ is the instance of the human module for modeling shapes of the source human, and $\mathcal{M}_{th}$ is another instance of the human module for generating novel target human shapes. 

Training consists of three stages: We first train $\mathcal{M}_{th}$ and $\mathbf{z}_{th}$ with $\mathbf{S}_{th}$, to leverage the wide variation of shapes and poses of samples in $\mathbf{S}_{th}$ for the multi-subject forward skinning module, $D_{th}$. For later stages, $D_{th}$ is used to initialize other deformation networks with its warping network $N_{th}$ frozen, to let all samples share the same canonical space. Next, we train $\mathcal{M}_{sh}$ and $\mathbf{z}_{sh}$ with $\mathbf{S}_{sh}$. For the last stage, using all the samples, we train $\mathcal{M}_{o}$, $\mathcal{M}_{comp}$, $\mathbf{z}_{sh}$, and $\mathbf{z}_{o}$ with the pre-trained $\mathcal{M}_{th}$, $\mathcal{M}_{sh}$ and $\mathbf{z}_{th}$ frozen. Note that $\mathbf{z}_{sh}$ for the last stage are re-initialized as the mean of $\mathbf{z}_{sh}$ after the second stage, denoted $\mathbf{\overline{z}}_{sh}$.
$\mathcal{M}_{comp}$ models all training samples using the feature vector from either $\mathcal{M}_{th}$ or $\mathcal{M}_{sh}$ for the human part, and from $\mathcal{M}_{o}$ for the object part. In the case of $\mathbf{S}_{th}$ and $\mathbf{S}_{sh}$ where scans are with no objects, we introduce a new latent code $\mathbf{z}_{emp}$ as an alternative input to $\mathcal{M}_{o}$ for no objects.
 
\noindent \textbf{Losses}: For the first stage, we use losses following~\cite{chen2022gdna}. We use the binary cross entropy loss $\mathcal{L}_{th}$ between the predicted occupancy of $\mathcal{M}_{th}$ and the ground truth occupancy. Note that $O^d(\cdot)$ and $F^d(\cdot)$ denote the occupancy field and SDF in posed space, respectively. We also use guidance losses $\mathcal{L}_{bone}$, $\mathcal{L}_{joint}$ and $\mathcal{L}_{warp}$ to aid training. $\mathcal{L}_{bone}$ encourages the occupancy of $\mathbf{x}_{bone}$ to be one, where $\mathbf{x}_{bone}$ are randomly selected points along the SMPL bones in canonical space. $\mathcal{L}_{joint}$ encourages the skinning weights of SMPL joints to be 0.5 for connected two bones and 0 for all other bones.  $\mathcal{L}_{warp}$ encourages deformation network $N$ to change body size consistently, by enforcing vertices of a fitted SMPL to warp to vertices of the mean SMPL shape, achieved by having shape parameter $\beta$ as zero. Lastly, we use $\mathcal{L}_{reg\_th}$ to regularize the latent code $\mathbf{z}_{th}$ to be close to zero.
\begin{gather}
    \mathcal{L}_{th} = BCE((O_{th}^d(\mathbf{x}^c, G_{th}(\mathbf{z}_{th})), o_{gt}) \\ 
    \mathcal{L}_{bone} = BCE((O_{th}(\mathbf{x}_{bone}, G_{th}(\mathbf{z}_{th})), 1) \\
    \mathcal{L}_{joint} = \lVert W(\mathbf{x}_{joint}, \mathbf{z}_{th}) - \mathbf{w}_{gt} \rVert \\
    \mathcal{L}_{warp} = \lVert N(\mathbf{v}(\beta), \beta) - \mathbf{v}(\beta_0) \rVert \\
    \mathcal{L}_{reg\_th} = \lVert \mathbf{z}_{th} \rVert
\end{gather}

For training the SDF network, we use L1 loss $\mathcal{L}_{sdf}$ between the predicted and the ground truth signed distance and L2 loss $\mathcal{L}_{nml}$ between the gradients of SDF and the ground truth normals of points on the surface. We additionally use $\mathcal{L}_{igr}$ for SDF to satisfy the Eikonal equation~\cite{gropp2020implicit} and $\mathcal{L}_{bbox}$ to prevent SDF values of off-surface points from being the zero-level surface as in~\cite{sitzmann2020implicit}.
 \begin{gather}
    \mathcal{L}_{sdf} = \lvert F_{th}^d(\mathbf{x^c}, G_{th}(\mathbf{z}_{th})) - d_{gt} \rvert \\
    \mathcal{L}_{nml} = \lVert \nabla F_{th}^d(\mathbf{x^c}, G_{th}(\mathbf{z}_{th})) - n_{gt} \rVert \\
    \mathcal{L}_{igr} = (\lVert \nabla F_{th}(\mathbf{x^c}, G_{th}(\mathbf{z}_{th})) \rVert - 1)^2 \\
    \mathcal{L}_{bbox} = exp(-\alpha \cdot \lvert F_{th}(\mathbf{x^c}, G_{th}(\mathbf{z}_{th})) \rvert), \alpha \gg 1
\end{gather}

For the second stage, we use the binary cross entropy loss $\mathcal{L}_{sh}$ between the predicted occupancy of $\mathcal{M}_{sh}$ and the ground truth occupancy, and $\mathcal{L}_{reg\_sh}$ to regularize the latent code $\mathbf{z}_{sh}$ to be close to zero. Since we initialize $D_{sh}$ with pre-trained $D_{th}$, additional guidance losses are not required. 

\begin{gather}
    \mathcal{L}_{sh} = BCE((O_{sh}^d(\mathbf{x}^c, G_{sh}(\mathbf{z}_{sh})), o_{gt}) \\
    \mathcal{L}_{reg\_sh} = \lVert \mathbf{z}_{sh} \rVert
\end{gather}

For the last stage, we use the binary cross entropy loss $\mathcal{L}_{comp}$ between the predicted occupancy of $\mathcal{M}_{comp}$ and the ground truth occupancy. We also use $\mathcal{L}_{o}$ between the predicted occupancy of $\mathcal{M}_{o}$ and the residual part of $\mathbf{S}_{sh+o}$ where $\mathcal{M}_{h}$ cannot explain. Moreover, we optimize $\mathbf{z}_{sh}$ by using the binary cross entropy loss $\mathcal{L}_{fit}$ between the output of $\mathcal{M}_{sh}$ and the ground truth occupancy. Finally, we regularize $\mathbf{z}_{sh}$ to be close to $\mathbf{\overline{z}}_{sh}$ and $\mathbf{z}_{o}$ to be close to zero.
\begin{gather}
    \mathcal{L}_{comp} = BCE((O_{comp}^d(\mathbf{x}^c, \mathbf{f}_h, \mathbf{f}_{o}), o_{gt}) \\ 
    \mathcal{L}_{o} = BCE((O_{o}(\mathbf{x}^c, G_{o}(\mathbf{z}_{o})), (1 - o_h) \cdot o_{comp}) \\ 
    \mathcal{L}_{fit} = BCE((O_{sh}^d(\mathbf{x}^c, G_{sh}(\mathbf{z}_{sh})), o_{gt}) \\ 
    \mathcal{L}_{reg\_sh} = \lVert \mathbf{z}_{sh} - \mathbf{\overline{z}}_{sh} \rVert \\
    \mathcal{L}_{reg\_o} = \lVert \mathbf{z}_{o} \rVert
\end{gather}

\section{Experiments}
\label{sec:experiments}
We evaluate our generative composition model across various scenarios. We first demonstrate the quality of the random 3D avatar creations from our model and the disentangled natures of human and object controls. 
Quantitative and qualitative comparisons against the previous SOTA~\cite{chen2022gdna} are performed, incorporating a user study via CloudResearch Connect. We also conduct ablation studies to validate our design choices. 

\subsection{Dataset}

\noindent \textbf{Our 3D Scans:} As described in Sec.~\ref{subsection:data_acq}, we use our multi-Kinect system to capture the source human with and without objects, $\mathbf{S}_{sh}$ (180 samples) and $\mathbf{S}_{sh+o}$ (342 samples). For $\mathbf{S}_{sh+o}$, we consider 4 categories of objects: 5 backpacks (77 samples in total), 6 outwear (94 samples), 8 scarves (89 samples), and 6 hats (82 samples). 

We run quantitative evaluation by focusing on backpacks as other objects such as outwear are already incorporated in $\mathbf{S}_{th}$. We use another set with 300 samples of the source human with backpacks only, denoted as $\mathbf{S}_{sh+bp}$. To build a testing set for FID computation in this quantitative evaluation, we further capture 343 samples of 3 different unseen identities who wear unseen backpacks. We denote this test dataset, $\mathbf{S}_{unseen+bp}$.

\noindent  \textbf{THuman2.0~\cite{tao2021function4d}:} THuman2.0\footnote{The THuman2.0 dataset was downloaded, accessed, and used in this research exclusively at SNU.} provides high-quality 3D dataset for dressed humans. We use 526 samples for $\mathbf{S}_{th}$.

\subsection{Qualitative Evaluation}
We demonstrate the expressive power and controllability of our composition model via inferences in various scenarios by controlling latent codes for humans $\mathbf{z}_{h}$ and object $\mathbf{z}_{o}$.

\begin{figure}[t]
\includegraphics[trim={1cm 1.5cm 1cm 0.2cm},clip,width=1.0\columnwidth]{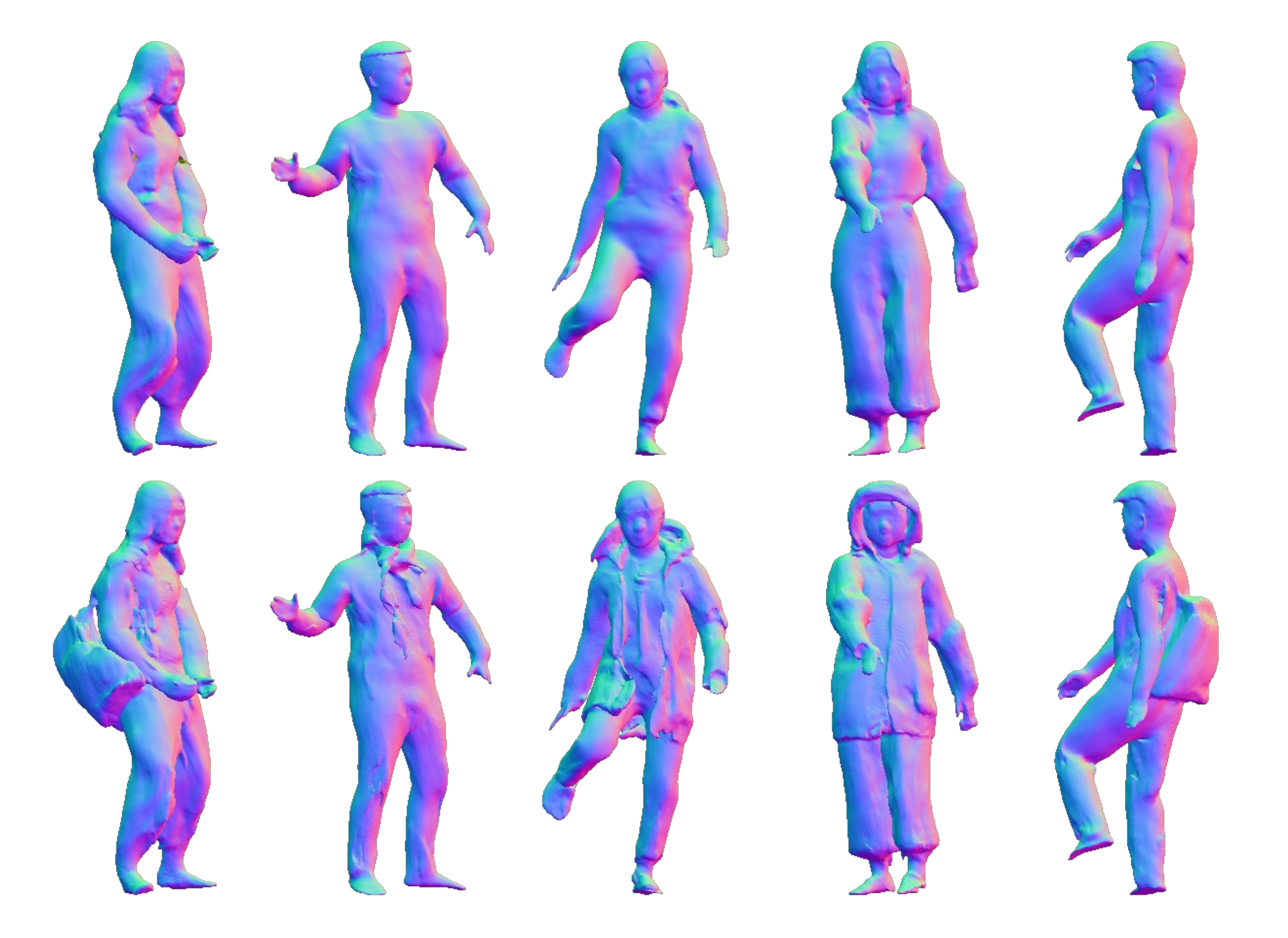}
\vspace{-15px}
\caption{\textbf{Random Generation.} Top row: randomly sampled outputs of the human module before composition. Bottom row: composition outputs of target humans on top with specific objects.}
\label{fig:generation}
\vspace{-5px}
\end{figure}
\noindent \textbf{Random Generation.} The 3D avatars created by attaching specific object latent codes $\mathbf{z}_{o}$ to random sampled human codes $\mathbf{z}_{h}$ are shown in Fig.~\ref{fig:generation} (bottom). The outputs of the human module $\mathcal{M}_{h}$ are also shown on the top of Fig. \ref{fig:generation} for reference. Our model enables the creation of diverse 3D avatars with controllable objects.

\begin{figure}[t]
\includegraphics[trim={0.3cm 1.1cm 0.3cm 0.2cm},clip,width=1.0\columnwidth]{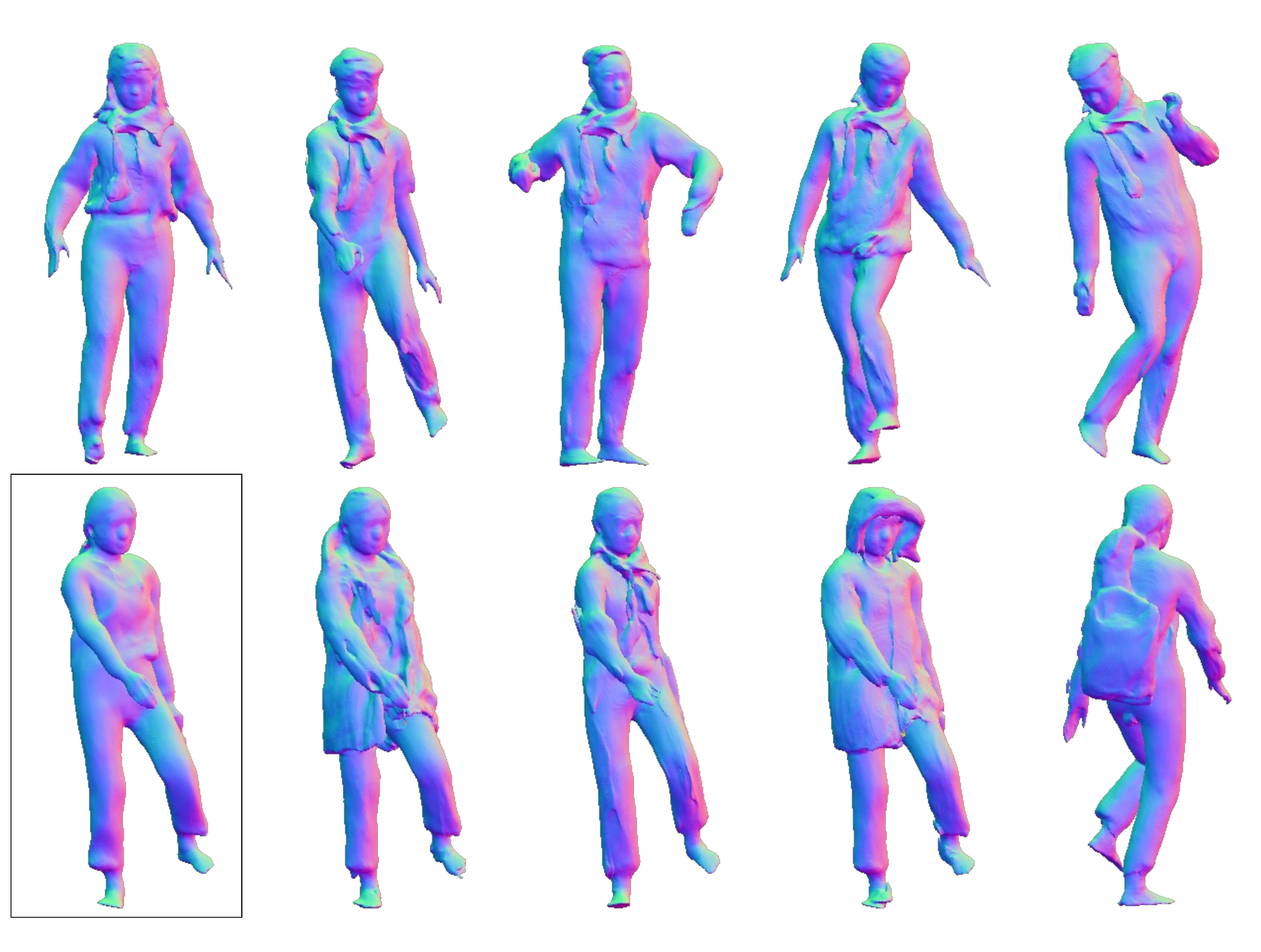}
\vspace{-18px}
\caption{\textbf{Disentangled Human and Object.} Top row: composition outputs of the same object (a scarf), added to different human identities. Bottom row: composition outputs of different objects added to the single human identity shown in the leftmost column.}
\label{fig:disentangled_hum_obj}
\vspace{-13px}
\end{figure}
\noindent \textbf{Disentangled Controls over Human and Objects.} 
To further test the disentangled nature of our composition model, we create 3D humans with objects by changing either human latent code or object latent code, as shown in Fig. \ref{fig:disentangled_hum_obj}. The examples on the top vary the human part by keeping the same object code that represents a scarf. On the bottom examples, we vary object codes for a fixed identity shown on the leftmost side. These results show the core advantage of our composition model in individual controls. 

\begin{figure}[t]
\includegraphics[trim={1cm 2cm 1cm 0.2cm},clip,width=1.0\columnwidth]{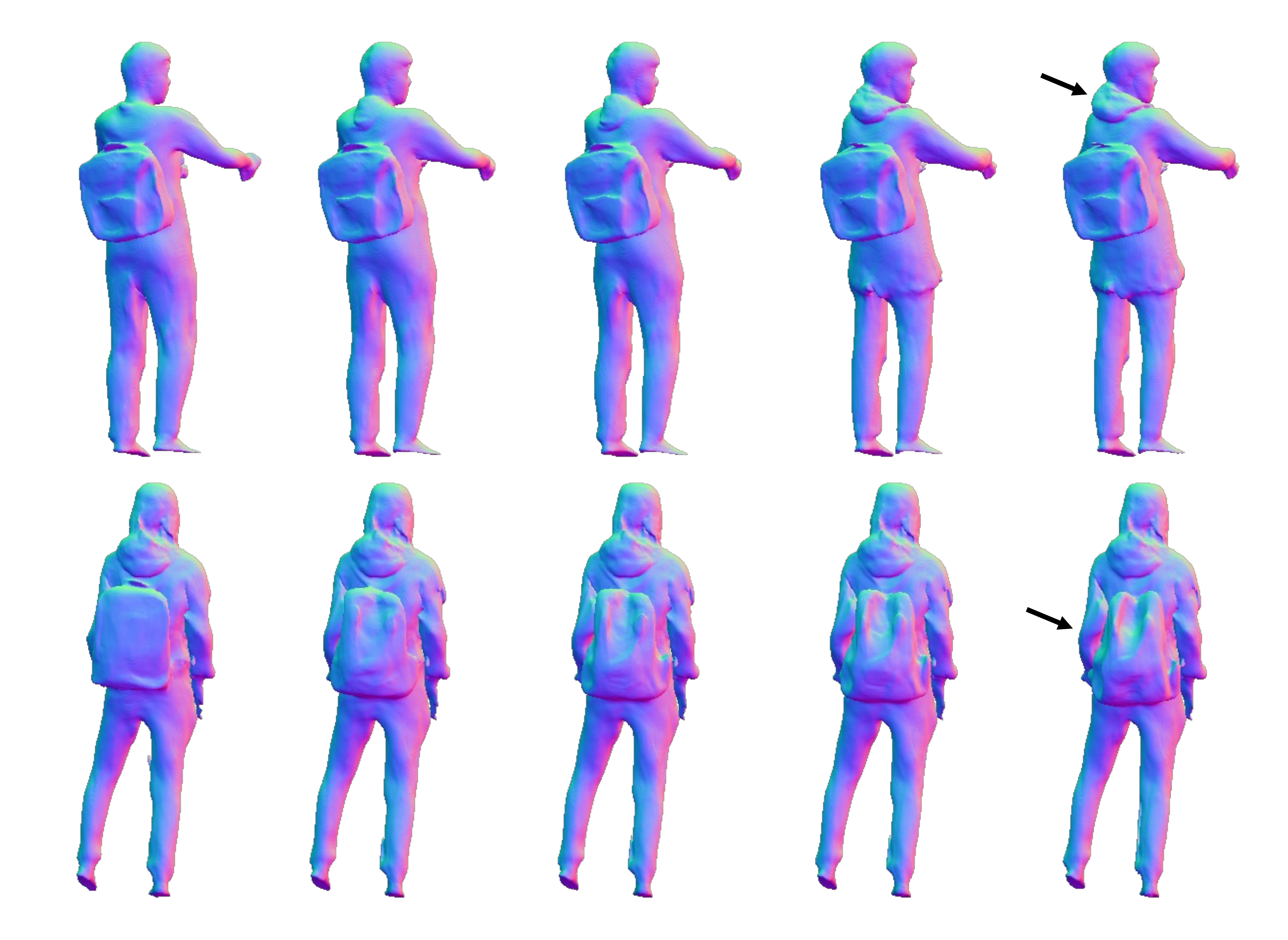}
\vspace{-16px}
\caption{\textbf{Interpolation.} Top row: human module interpolation. Bottom row: object module interpolation. Notice that interpolating one module doesn't deteriorate the geometry of the other.}
\label{fig:interpolation}
\vspace{-15px}
\end{figure}
\noindent \textbf{Interpolation.} 
Fig. \ref{fig:interpolation} demonstrates smooth interpolation of each module without deteriorating the other module. 

\begin{figure}[t]
\includegraphics[trim={1cm 1cm 1cm 0.2cm},clip,width=1.0\columnwidth]{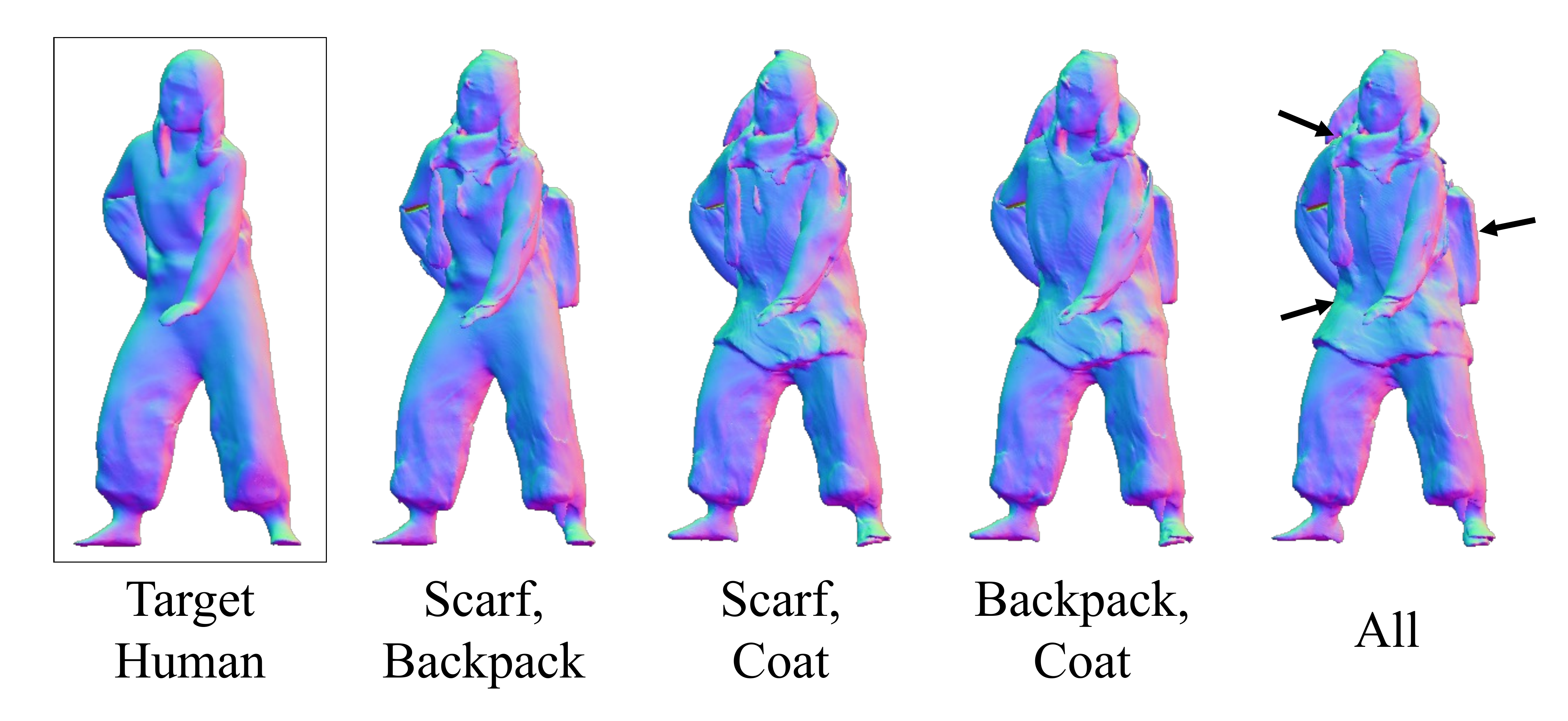}
\vspace{-18px}
\caption{\textbf{Composition of Multiple Objects.} Two or more objects are added to the leftmost human. Note that our train data contain no scans of the source human with multiple objects.}
\vspace{-5px}
\label{fig:multiple_objects}
\end{figure}
\noindent \textbf{Composition of Multiple Objects.} Fig. \ref{fig:multiple_objects} shows that our system allows the composition of multiple objects. To add multiple objects, we use the latent code of each object and get the occupancy and the feature vector of objects. Using the normalized occupancy of multiple objects as weights, we calculate the weighted sum of feature vectors. The aggregated feature is then fed to the composition module along with the human feature to get the final composition output.
Note that our dataset has no such sample with multiple objects.

\begin{table}[t]
\centering
\small{

\begin{tabular}{lcc}
\toprule
Method & FID & User Preference \\
\midrule
gDNA (w/ object) & 41.71 & 43.6\% \\
Arith. gDNA (w/ object) & 73.81 & 13.6\% \\
Ours (Naive composition) & 55.29 & 22.4\% \\
Ours & 51.03 & 100\% - (above) \\
\bottomrule
\end{tabular}
}
\vspace{-5px}
\caption{Quantitative evaluation of the importance of compositional modeling. User preference score reflects the frequency with which participants of our perceptual study favored each method over ours.}


\vspace{-15px}
\label{tab:ablation}
\end{table}

\begin{figure}[t]
\includegraphics[trim={1cm 1cm 1cm 0.2cm},clip,width=1.0\columnwidth]{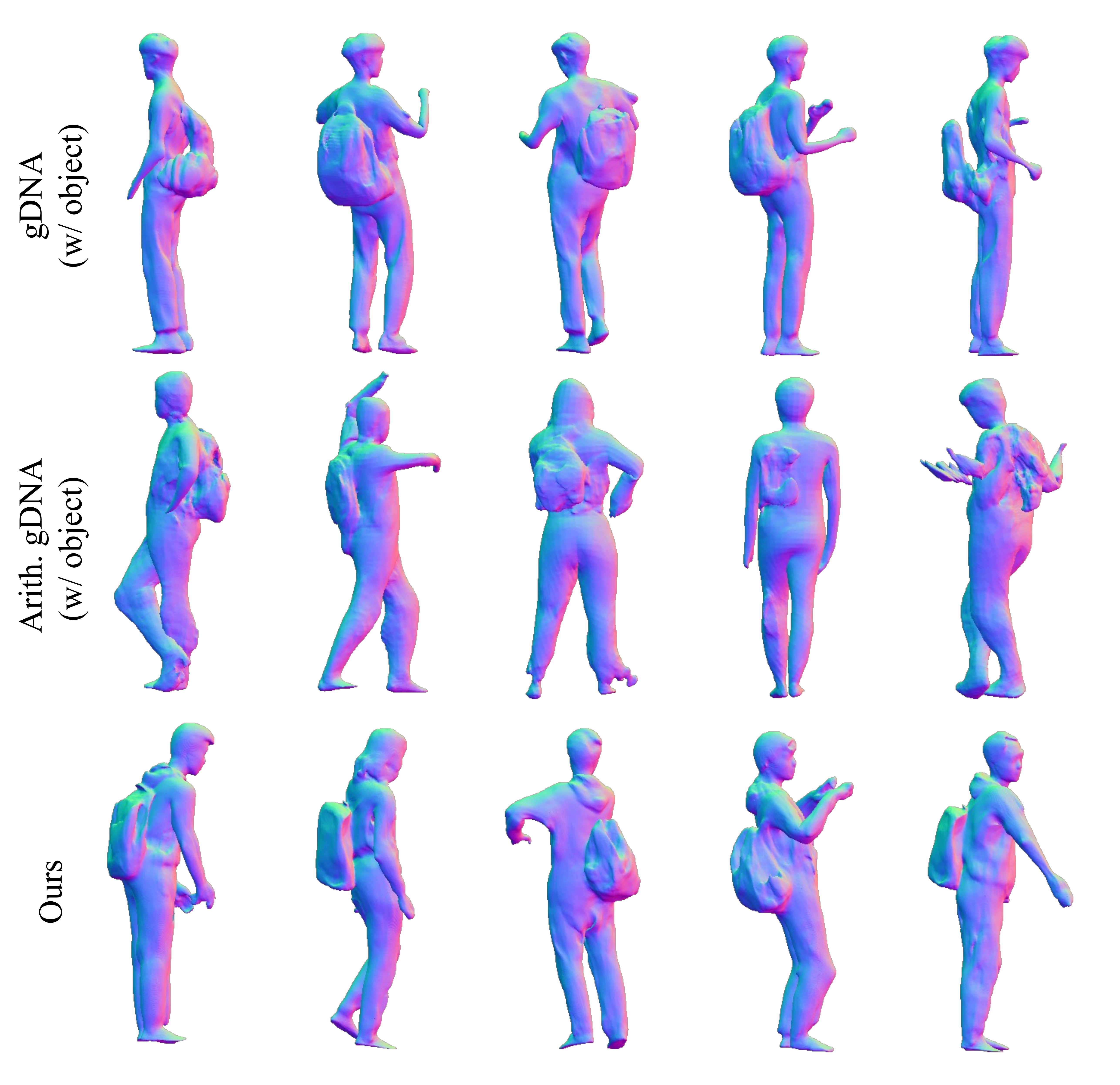}
\vspace{-15px}
\caption{\textbf{Qualitative Comparison on Compositional Modeling.} Compared to our method, baselines suffer from generating outputs of diverse humans with complete objects.}
\vspace{-10px}
\label{fig:ablation1}
\end{figure}

\subsection{Comparison with SOTA}
\vspace{-6px}
Since our method is the first generative model for compositing humans and objects, there is no direct competitor, and comparison with the previous non-compositional model such as gDNA is non-trivial. 
To make the assessment possible at our best, we consider a specific scenario where a user wants to create samples with a specific object category, being the backpack here. To provide such controllability on gDNA, we first extend the gDNA model with our dataset. Note that, in this evaluation, we use the same dataset  $\mathbf{S}_{th}$, $\mathbf{S}_{sh}$, $\mathbf{S}_{sh+bp}$ for training both our model and gDNA.

\noindent \textbf{Extending gDNA for Composition.}
We train gDNA model using the public code with our datasets. Both human-only outputs and the ones with a backpack can be sampled from the trained model. To intentionally generate outputs with a backpack, we search the latent codes associated with the training samples with backpacks and fit a gaussian from which we can perform a sampling. We denote this baseline method as \emph{`gDNA (w/ object)'}.

The second possible extension of gDNA is based on the arithmetic operation among gDNA's latent codes, which is widely used for GAN-based image manipulation~\cite{radford2015unsupervised}. We found that gDNA's original framework allows some level of composition by adding or subtracting the latent codes. Specifically, we choose a latent code $\mathbf{z}_{sh}^*$ for the source human without a backpack and another latent code $\mathbf{z}_{sh+bp}^*$ for the source human with the backpack. We simply take their subtraction $\mathbf{z}_{bp} = \mathbf{z}_{sh+bp}^* - \mathbf{z}_{sh}^*$, which can be considered as a residual for the backpack. We found that composition can be performed by adding this residual to another human's latent code, that is $\mathbf{z}_{bp} + \mathbf{z}_{th}$. We denote this baseline method as \emph{`Arith. gDNA (w/ object)'}.

\noindent \textbf{Qualitative Comparison with User Study.}
The visual comparison between ours and the extended gDNAs is shown in Fig.~\ref{fig:ablation1}.
In the first row, we show random samples generated from \emph{`gDNA (w/ object)'}. Since the human scans with the backpack are only of the source human's (other samples from $\mathbf{S}_{th}$ do not have any backpack), the generated outputs lack shape variety for the human part, producing always the source human's identity.
In the second row of Fig. \ref{fig:ablation1}, backpacks are added to novel identities; however, the method suffers from lack of details on both humans and objects.
In contrast, the outputs of our method shown in the last row show strong generalization by creating diverse human identities with naturally attached detailed objects.

To further validate this comparison, we perform a user test (A/B test) on  CloudResearch Connect. We render samples from three viewpoints (same views for all) and show ours with each baseline (A/B examples) in a random order to each subject. Each subject answers 5 questions per baseline by choosing more authentic 3D human samples. The data was collected from 50 subjects.
The results are shown in the ``User Preference'' column in Tab.~\ref{tab:ablation}. As shown, our methods are preferred over extended gDNA baselines.
Moreover, to confirm the diversity of identities in our method and \emph{`gDNA (w/ object)'}, 50 subjects were shown the rendering of the source human and were asked to choose samples that don't resemble the source human. Samples of our method were chosen by 92.4\%, indicating that \emph{`gDNA (w/ object)'} suffers to generate novel identities with a backpack.

\noindent \textbf{Quantiative Evaluation via FID.}
To evaluate the generation quality of our method, we compare Fr\'{e}chet Inception Distance (FID) between the 2D normal renderings of the test dataset $\mathbf{S}_{unseen+bp}$ and the generated outputs, following \cite{chen2022gdna}.
The result is shown in Tab.~\ref{tab:ablation}. \emph{`gDNA (w/ object)'} has a relatively better score than ours, due to the fact that it only samples 3D humans around $\mathbf{S}_{sh+bp}$, which are always close to the GT samples.
A more fair comparison is between ours and \emph{`Arith. gDNA (w/ object)'}, where both approaches try to attach the backpack to novel identities. Our method significantly outperforms this baseline.

\begin{table}[t]

\centering
\small{
\begin{tabular}{lcc}
\toprule
Method & Pred-to-Scan${\downarrow}$ & Scan-to-Pred${\downarrow}$   \\
\midrule
gDNA & 0.0162 & 0.0190 \\
gDNA(w/ object) & 0.0218 & 0.0112 \\
Ours & \textbf{0.0116} & \textbf{0.0099} \\
\bottomrule
\end{tabular}
}
\vspace{-5px}
\caption{Fitting accuracy comparison with the SOTA method~\cite{chen2022gdna}.}
\vspace{-5px}
\label{tab:fitting}
\end{table}

\begin{figure}[t]
\includegraphics[trim={1cm 1.2cm 1cm 0.2cm},clip,width=1.0\columnwidth]{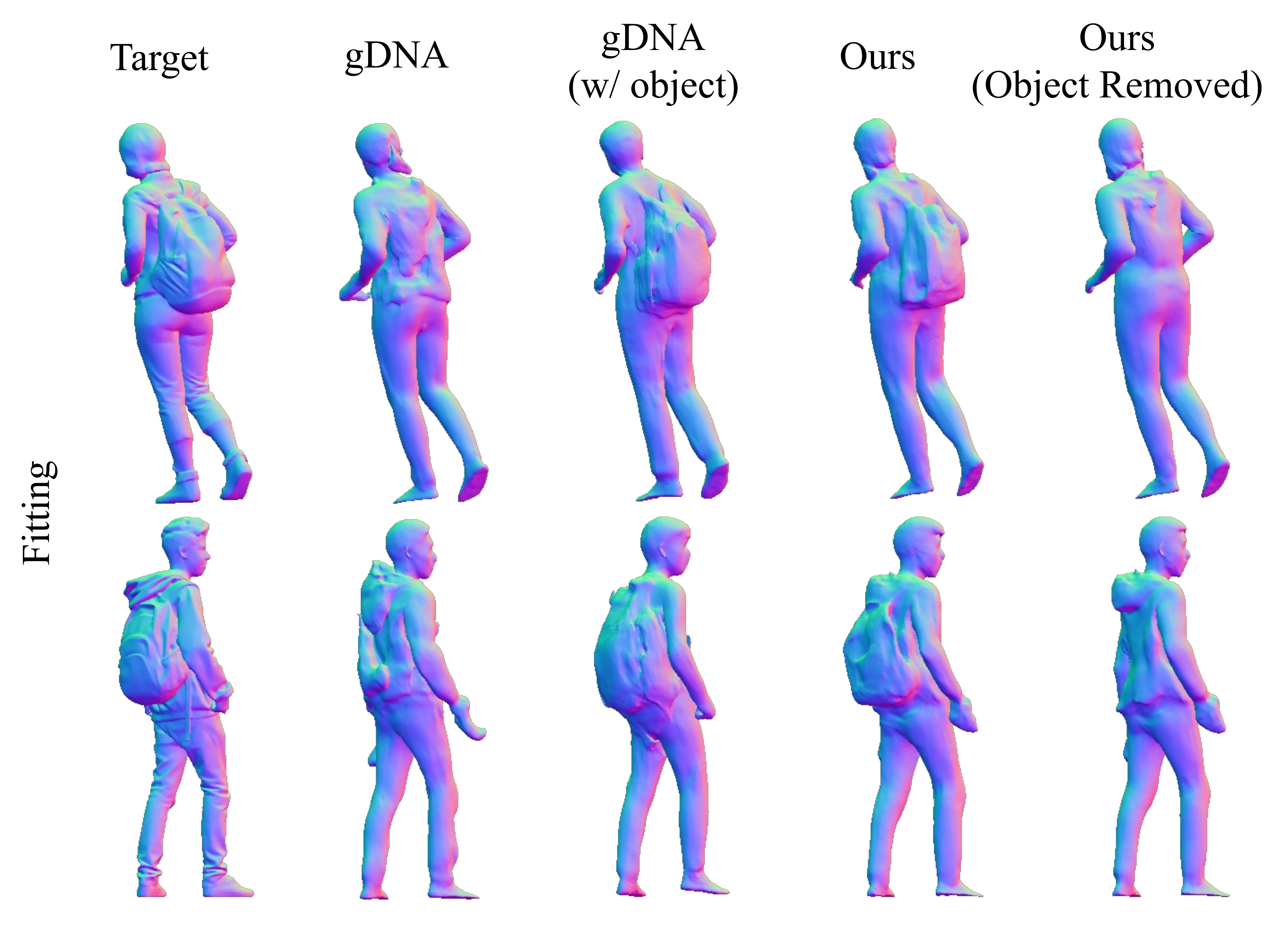}
\vspace{-20px}
\caption{\textbf{Fitting and Object Removal.} Compared to baselines, our method successfully explains both human shapes and object shapes, enabling the natural removal of objects after fitting.}
\vspace{-12px}
\label{fig:fitting}
\end{figure}

\noindent \textbf{Performance on Fitting.}
\label{subsec:result_fitting}
We evaluate the expressiveness of our model by fitting it to unseen scans with objects. As a baseline, we consider gDNA~\cite{chen2022gdna} as it demonstrates better fitting results on 3D clothed human scans over other SOTA methods~\cite{palafox2021npms, corona2021smplicit}. Besides the original gDNA trained with $\mathbf{S}_{th}$, we also consider gDNA trained with $\mathbf{S}_{th}$, $\mathbf{S}_{sh}$ and $\mathbf{S}_{sh+bp}$ (\emph{`gDNA (w/ object)'}) to enable fitting of the object part. We use scans with backpacks from Renderpeople\footnote{RenderPeople was downloaded, accessed, and used in this research exclusively at SNU.}~\cite{renderpeople} and captured dataset $\mathbf{S}_{unseen+bp}$ for fitting comparison.

As shown in Tab. \ref{tab:fitting}, our method reports better fitting accuracy than the baselines. Our method effectively fits the geometry of both humans and objects while baselines only reconstruct either the human part or the object part as shown in Fig. \ref{fig:fitting}. Moreover, since our method separately models humans and objects, it enables the high-quality removal of objects after fitting.

\vspace{-6px}
\subsection{Ablation Study}

\begin{figure}[t]
\includegraphics[trim={1cm 1.2cm 1cm 0.4cm},clip,width=1.0\columnwidth]{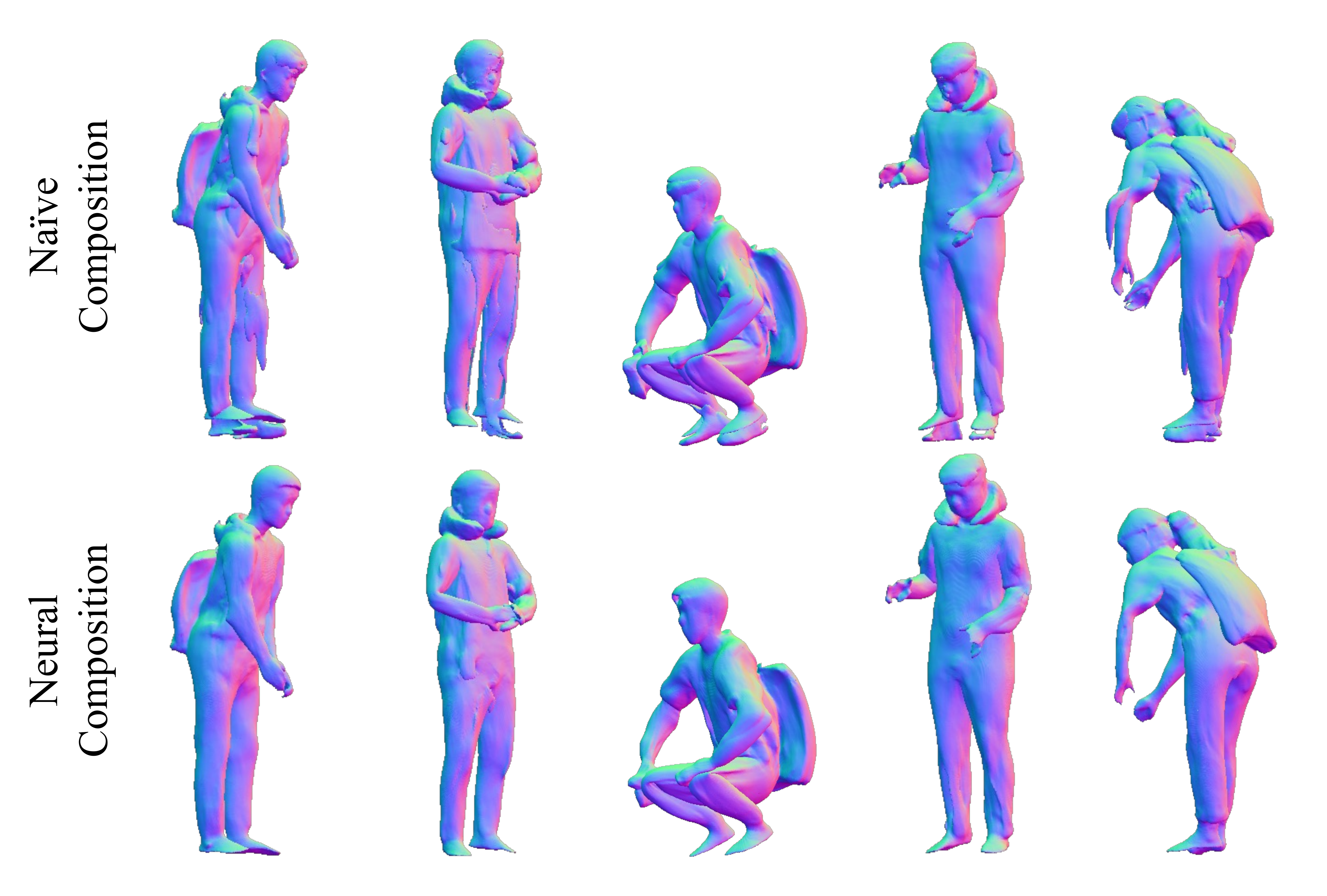}
\vspace{-20px}
\caption{\textbf{Composition Comparison.} While naive composition suffers from severe artifacts, neural composition reduces these artifacts and produces high-quality outputs.}
\vspace{-18px}
\label{fig:ablation2}
\end{figure}

\vspace{-6px}
\noindent \textbf{Neural Composition.} Our system provides two ways of extracting the final composition output. One is by using $o_{comp}$: neural composition, and the other is by using the maximum value between $o_{h}$ and $o_{o}$ of queried points: naive composition. We verify the necessity of using neural composition in order to generate high-quality outputs of humans with objects. Compared to naive composition, neural composition remarkably reduces the artifacts induced by the imperfect fitting of the source human, resulting in lower FID values (Tab. \ref{tab:ablation}). Qualitative comparison is presented in Fig. \ref{fig:ablation2}.

\vspace{-8px}
\section{Discussion}
\vspace{-6px}
\label{sec:conclusion}

We present a novel framework for learning a compositional generative model of humans and objects. 
Our compositional generative model provides separate control over the human part and the object part. 
To train our compositional model without manual annotation for the object geometries, we propose to leverage 3D scans of a single person with and without objects. 
Our results show that the learned generative model for the object part can be authentically transferred to novel human identities.

\noindent\textbf{Limitations and Future Work.} While our approach is general and supports diverse objects, decomposing thin layers of clothing in an unsupervised manner remains a challenge due to the limited precision of 3D scans. Extending our approach to modeling from RGB images is also an exciting research direction for future work.

{\footnotesize\noindent\textbf{Acknowledgements:} This work of H. Joo and T. Kim was supported by SNU Creative-Pioneering Researchers Program and IITP grant funded by the Korean government (MSIT) [NO.2021-0-01343 and No.2022-0-00156]}

{\small
\bibliographystyle{ieee_fullname}
\bibliography{11_references}
}

\ifarxiv \clearpage 

\appendix
\renewcommand{\thesection}{\Alph{section}}
\setcounter{page}{1}
\setcounter{section}{0}

\section{Implementation details}    
\subsection{Network Architectures}
Latent codes assigned to each scan, $\mathbf{z}_{th}$, $\mathbf{z}_{sh}$, and $\mathbf{z}_{o}$ are 64-dimensional. For $\mathbf{z}_{o}$, we use its first 5 bits to encode the object category via one-hot encoding and optimize only the last 59 bits during training. 
The generator $G$ of the human module and the object module generates the $256 \times 256 \times 64$ feature image from a constant vector of size $256 \times 16 \times 16$ via 4 layers of (bilinear upsampler with a scale factor of 2, 2D convolution of kernel size 3 and stride 1, adaIN for conditioning the generator with the latent code $\mathbf{z}$, and leaky ReLU activations). The $256 \times 256 \times 64$ output feature image is split into one $256 \times 256 \times 32$ and two $256 \times 128 \times 32$ to form a tri-plane feature map. Note that the feature map is 128-dimensional along z-axis and 256-dimensional along other axes.
The decoder for predicting the occupancy of the human module and the object module is a multi-layer perceptron having the intermediate neuron size of (256, 256, 256, 229, 1) with skip connection from the input features to the 4th layer and nonlinear activations of softplus with $\beta = 100$ except for the last layer that uses sigmoid. As an input, it takes the Cartesian coordinates in canonical space which are encoded using a positional encoding with 4 frequency components, and the 32-dimensional feature queried from the generated tri-plane. 
The decoder for predicting SDF of the human module has the same architecture as the decoder for predicting the occupancy, except that it has no activations for the last layer.
The decoder for predicting the occupancy of the composition module has the same architecture as the decoders for predicting the occupancy of other modules. However, instead of taking in the feature from the generated tri-plane as an input, it takes in the intermediate latent feature vectors before the last layer of the decoders for predicting the occupancy of the human module and object module, which are 229-dimensional each.

Our deformation networks $D = (W, N)$ follow the architecture of the deformer of gDNA~\cite{chen2022gdna}. The skinning network $W$ is a multi-layer perceptron having the intermediate neuron size of (128, 128, 128, 128, 24) with nonlinear activations of softplus with $\beta = 100$, except for the last layer that uses softmax in order to get normalized skinning weights. As an input, it takes the Cartesian coordinates in canonical space and the latent code $\mathbf{z} \in \mathbb{R}^{64}$ of the training sample. The warping network $N$ is also a multi-layer perceptron having the intermediate neuron size of (128, 128, 128, 128, 3) with nonlinear activations of softplus. As an input, it takes the Cartesian coordinates in canonical space and the SMPL shape parameter $\beta \in \mathbb{R}^{10}$ of the training sample. The input Cartesian coordinates are passed to the last layer for the network to learn residual displacements.

\subsection{Training Procedure}

Our training consists of three stages. First, we train $\mathcal{M}_{th}$ and $\mathbf{z}_{th}$ with $\mathbf{S}_{th}$ with losses following ~\cite{chen2021snarf, chen2022gdna} and additional losses to train the SDF network. The total loss $\mathcal{L}_{M_{th}}$ is as follows:
\begin{gather}
    \mathcal{L}_{M_{th}} = \mathcal{L}_{th} + \lambda_{bone}\mathcal{L}_{bone} + \lambda_{joint}\mathcal{L}_{joint} + \lambda_{warp}\mathcal{L}_{warp} \\ \nonumber
    + \lambda_{reg\_th}\mathcal{L}_{reg\_th} + \mathcal{L}_{sdf} + \mathcal{L}_{nml} + \mathcal{L}_{igr} + \mathcal{L}_{bbox},
\end{gather}
where $\lambda_{warp} = 10$ and $\lambda_{reg\_th} = 10^{-3}$. We set $\lambda_{bone} = 1$ and $\lambda_{joint} = 10$ only for the first epoch and 0 afterwards.

For the second stage, we train $\mathcal{M}_{sh}$ and $\mathbf{z}_{sh}$ with $\mathbf{S}_{sh}$ with the total loss $\mathcal{L}_{M_{th}}$ being,
\begin{gather}
    \mathcal{L}_{M_{sh}} = \mathcal{L}_{sh} + \lambda_{reg\_sh}\mathcal{L}_{reg\_sh},
\end{gather}
where $\lambda_{reg\_sh} = 10^{-3}$. As described in the main paper, since we initialize $D_{sh}$ with
the pre-trained $D_{th}$, additional guidance losses as in the first stage are not required. Note that since it is not our primary objective to model the detailed surface of the source human, we don't utilize the hybrid modeling of occupancy and SDF for $\mathcal{M}_{sh}$.

For the last stage, we train $\mathcal{M}_{o}$, $\mathcal{M}_{comp}$, $\mathbf{z}_{sh}$, and $\mathbf{z}_{o}$ with the pre-trained $\mathcal{M}_{th}$, $\mathcal{M}_{sh}$ and $\mathbf{z}_{th}$ frozen. As described in the main paper, $\mathbf{z}_{sh}$ for the last stage are re-initialized as the mean of $\mathbf{z}_{sh}$
after the second stage. The total loss $\mathcal{L}$ is as follows:
\begin{gather}
    \mathcal{L} = \mathcal{L}_{comp} + \mathcal{L}_{o} + \lambda_{fit}\mathcal{L}_{fit} \\ \nonumber
    + \lambda_{reg\_sh}\mathcal{L}_{reg\_sh} + \lambda_{reg\_o}\mathcal{L}_{reg\_o},
\end{gather}
where $\lambda_{fit} = 0.2$, $\lambda_{reg\_sh} = 50$, and $\lambda_{reg\_sh} = 10^{-3}$.

We train each stage with the Adam optimizer with a learning rate of 0.001 without decay. All stages are trained for 300 epochs.

\subsection{Inference}

We generate the composited canonical shapes of general people with objects by random sampling $\mathbf{z}_{th}$ and $\mathbf{z}_{o}$ from the Gaussian distribution fitted to each set of latent codes. We then extract meshes using $o_{comp}$ with a resolution of $256^3$. We finally repose the output mesh using the SMPL pose parameter with the learned skinning fields.

\section{Data}

\subsection{Acquisition}

We collect 3D scans of the source human with and without objects using a system with synchronized and calibrated 8 Azure Kinects. We capture data 5FPS with the resolution of $2048 \times 1536$ for the RGB cameras, and $1024 \times 1024$ for the depth cameras. We perform image-based calibration using COLMAP~\cite{schonberger2016structure} and adjust the optimized camera extrinsics to real-world scale based on the corresponding depth maps. 
We apply KinectFusion~\cite{newcombe2011kinectfusion} with the code from the repository~\footnote{\url{https://github.com/andyzeng/tsdf-fusion-python}} to fuse the captured depth maps with the voxel resolution of 1.5mm. We reconstruct watertight meshes from the fused output using screened-poisson surface reconstruction~\cite{kazhdan2013screened} of depth 9.
In order to obtain SMPL parameters for each captured scan, we use the multi-view extension of SMPLify~\cite{Bogo2016} with the code from the repository~\footnote{\url{https://github.com/ZhengZerong/MultiviewSMPLifyX}}.
For each scan, we render images from 18 viewpoints and detect 2D keypoints using OpenPose~\cite{cao2017realtime}, and apply the multi-view extension of SMPLify to estimate SMPL parameters for each scan.

\subsection{Data Statistics}

We use 180 samples for $\mathbf{S}_{sh}$ and 342 samples for $\mathbf{S}_{sh+o}$. For $\mathbf{S}_{sh+o}$, we consider 4 categories of objects: 5 backpacks (77 samples in total), 6 outwear (94 samples), 8 scarves (89 samples), and 6 hats (82 samples). 
For running the quantitative evaluation focused on backpacks, we use another set with 300 samples of the source human with 5 backpacks, denoted as $\mathbf{S}_{sh+bp}$. 
To build a testing set for FID computation, we further capture 343 samples of 3 different unseen identities who wear unseen backpacks, denoted as $\mathbf{S}_{unseen+bp}$.
We also use 526 samples of THuman2.0~\cite{tao2021function4d} for $\mathbf{S}_{th}$.

\section{Discussion}

\begin{table}[t]

\centering
\begin{tabular}{lccc}
\toprule
Method & Chamfer${\downarrow}$ & P2S${\downarrow}$ & Normal${\downarrow}$  \\
\midrule
Occ & 0.0140 & 0.0169 & 0.0092 \\
Occ \& SDF & \textbf{0.0098} & \textbf{0.0128} & \textbf{0.0074} \\
\bottomrule
\end{tabular}
\caption{Quantitative evaluation of the significance of using the hybrid modeling of occupancy and SDF is presented.}
\label{tab:occ_sdf}
\end{table}

\begin{figure}[t]
\includegraphics[trim={1cm 1cm 1cm 0},clip,width=1.0\columnwidth]{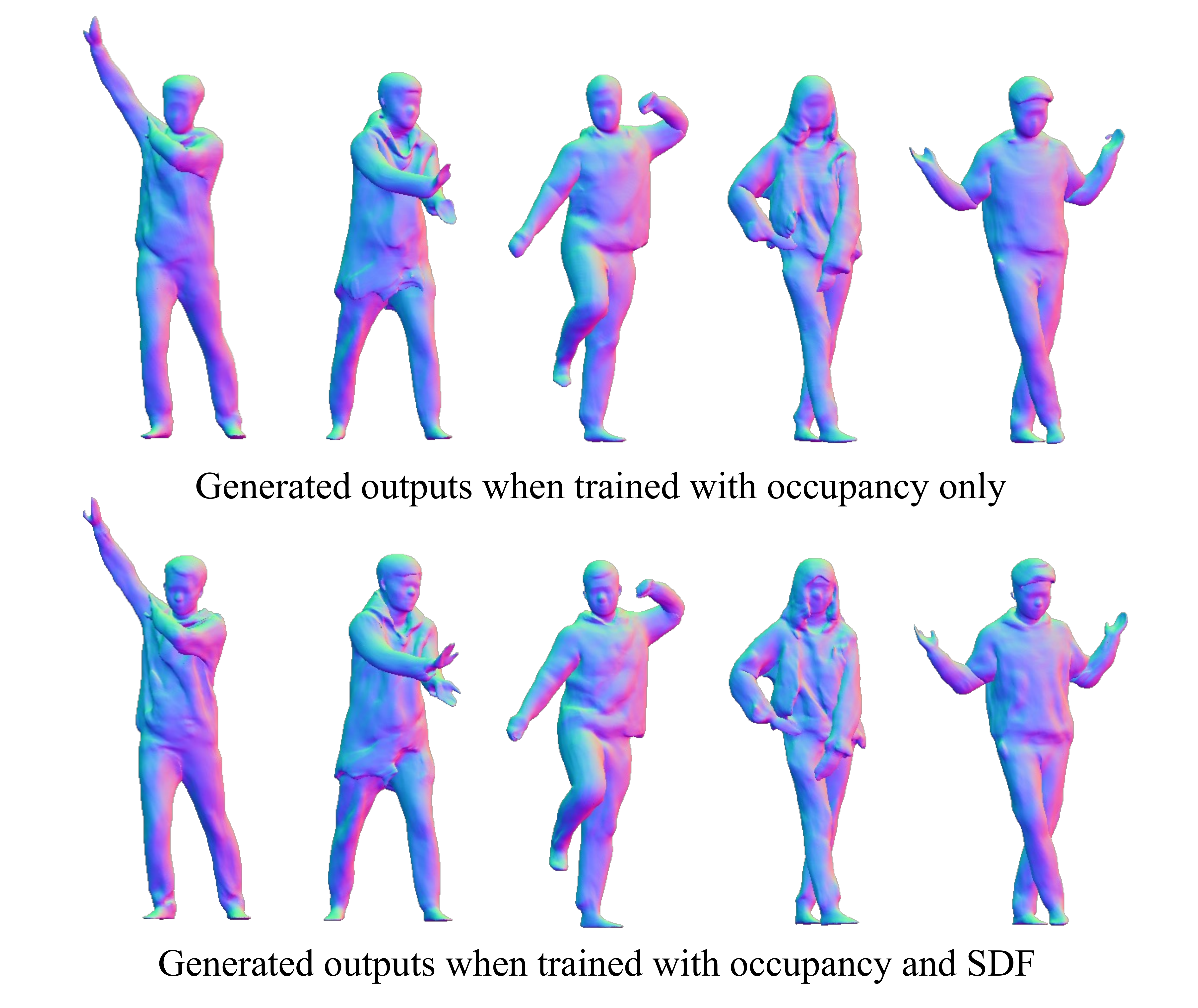}
\caption{\textbf{Qualitative Comparison on Introducing SDF Network in the Human Module.} Top row: Generated outputs when trained with occupancy only. Bottom row: Generated outputs when trained with the hybrid modeling of occupancy and SDF. Additionally predicting the SDF improves the details of generated outputs.}
\label{fig:ablation3}
\end{figure}

\subsection{Geometry Modeling with SDF}
As mentioned in the main paper, we model detailed geometry by jointly predicting SDF together with the occupancy fields. We find that directly replacing the occupancy with the SDF leads to failures in canonicalization. Among the set of correspondences resulting from multiple initials for the root finding algorithm, previous work that uses occupancy representation~\cite{chen2021snarf,chen2022gdna} determines the final correspondence by choosing the point with the highest estimated occupancy. However, in the case of the SDF representation, we empirically find out that choosing the point by only utilizing the estimated SDF leads to poor canonicalization. Moreover, using a single initial by linearly combining the skinning weights of the nearest neighbor on the fitted SMPL mesh and the inverse bone transformations as in~\cite{wang2022arah, jiang2022selfrecon} also leads to incorrect canonicalization. Hence, we utilize a hybrid modeling of occupancy and SDF by leveraging the advantage of each representation. While directly supervising SDF on the surface normals, we select final correspondences and train the deformation networks using occupancy. For stable training, it is crucial to disable the backpropagation of gradients from the SDF head to the deformation networks and let only the occupancy head supervise them.

We verify the significance of predicting both occupancy and SDF over predicting only occupancy to generate outputs with higher frequency details. For each method, we reconstruct the ground truth data used for training with assigned latent codes. We compute the Chamfer distance and point-to-surface distance (P2S) between the ground truth and the reconstruction output. We also render 2D normal maps from fixed views and compute the L2 error (Normal). As demonstrated in Tab. \ref{tab:occ_sdf}, reconstruction outputs are improved when both occupancy and SDF are predicted. In Fig. \ref{fig:ablation3} we show the qualitative comparison between samples generated via each method.

\section{Quantitative Evaluation Details}

\subsection{FID Computation}
We compute FID score using the code from the repository\footnote{\url{https://github.com/mseitzer/pytorch-fid}}. For the test set, we render 2D normal maps in resolution $256^2$ of 343 samples in $\mathbf{S}_{unseen+bp}$ from 18 viewpoints, resulting in 6174 images. For each method, we generate 200 samples in random body sizes and poses of the $\mathbf{S}_{unseen+bp}$ and similarly render 2D normal maps in resolution $256^2$ from 18 viewpoints, resulting in 3600 images.

\begin{figure}[t]
\includegraphics[trim={1cm 1cm 1cm 0},clip,width=1.0\columnwidth]{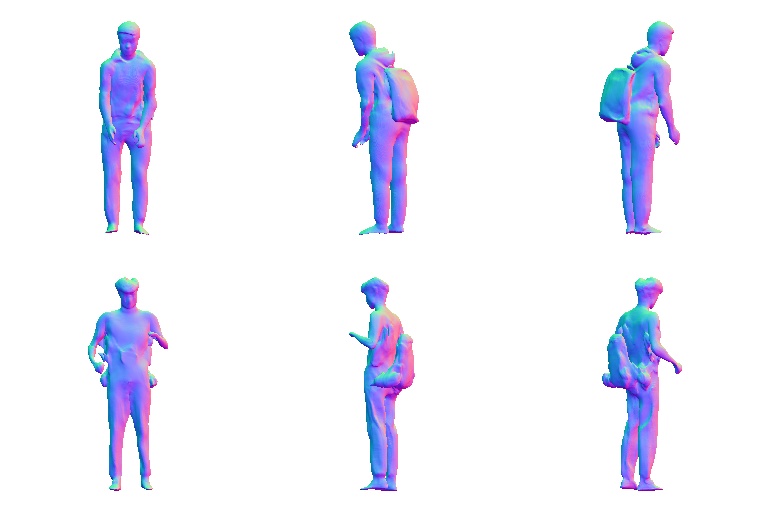}
\caption{\textbf{Example Images of the First User Study.} Subjects are asked to choose the sample with a more authentic shape between top and bottom.}
\label{fig:user_study_1}
\end{figure}

\begin{figure}[t]
\includegraphics[trim={1cm 1cm 1cm 0},clip,width=1.0\columnwidth]{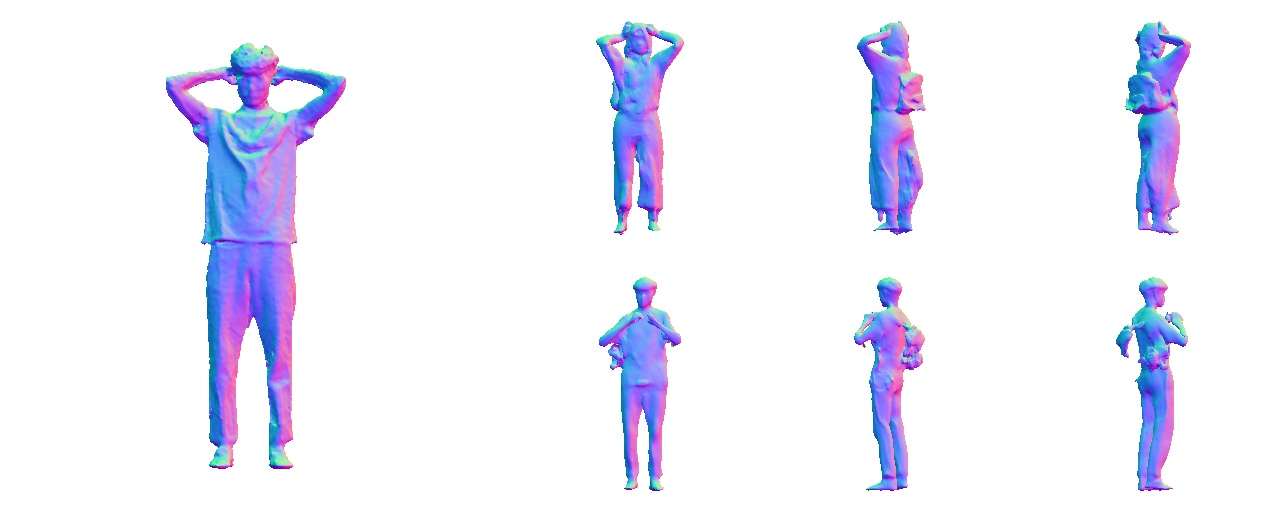}
\caption{\textbf{Example Images of the Second User Study.} Subjects are asked to choose the sample that does not resemble the shape of the source human shown on the left, between top and bottom.}
\label{fig:user_study_2}
\end{figure}

\subsection{User Preference Study}
We perform two user preference studies (A/B test) via CloudResearch Connect. The first study aims to validate the generation quality of our method over all baselines, and the second study aims to validate the generation diversity of our method over \emph{`gDNA (w/ object)'}. 

For the first user study, we show a sample generated with our method along with another sample generated with one of the baseline methods in random order. For each sample, we render 2D normal maps in resolution $256^2$ from 3 viewpoints. We ask 50 subjects to answer 5 A/B pairs per baseline by choosing the preferred sample with a more authentic shape. An example of a question is presented in Fig.~\ref{fig:user_study_1}

For the second user study, we only compare our method with the baseline, \emph{`gDNA (w/ object)'}, with a different protocol. In this study, we similarly render the normal maps from 3 viewpoints from each method and additionally show an image of the source human along with the A/B pairs. Then, we request the observers to choose the sample that looks more different from the source human. The test is intended to see whether the methods can produce diverse human identities with objects, sufficiently different from the source human's appearance. An example of a question is presented in Fig.~\ref{fig:user_study_2}. Similar to the first study, we ask 50 subjects to answer 5 A/B pairs by choosing the sample that better satisfies the question.

\subsection{Fitting Comparison}
For fitting our model to unseen scans with objects, we follow the fitting process of gDNA~\cite{chen2022gdna}. During fitting, we optimize the latent code for the human part, $\mathbf{z}_{h}$, the latent code for the object part, $\mathbf{z}_{o}$, and the SMPL shape parameter $\beta$ with other network frozen. We use $\mathcal{M}_{th}$ for the human module. We initialize $\mathbf{z}_{h}$ and $\mathbf{z}_{o}$ each with 8 randomly sampled codes from the Gaussian distribution fitted to each set of latent codes. $\beta$ is initialized with the obtained SMPL shape parameter during our data acquisition process.
The loss $\mathcal{L}_{fitting}$ used for fitting raw scans is as follows:
\begin{gather}
    \mathcal{L}_{fitting} = \mathcal{L}_{comp} + \lambda_{reg\_h}\mathcal{L}_{reg\_h} + \lambda_{reg\_o}\mathcal{L}_{reg\_o} \\ 
    \mathcal{L}_{comp} = BCE(o_{comp}, o_{unseen}) \\
    \mathcal{L}_{reg\_h} = \lVert \mathbf{z}_{h} \rVert \\
    \mathcal{L}_{reg\_o} = \lVert \mathbf{z}_{o} \rVert,
\end{gather}
where $\lambda_{reg\_h} = 50$ and $\lambda_{reg\_o} = 50$. We optimize for 500 iterations using the Adam optimizer with a learning rate of 0.01 without any weight decay or learning rate decay. Of 8 fitted outputs, the one with the minimum bi-directional Chamfer distance to the target scan is chosen as the final output.

\section{Addtional Qualitative Results}

Please refer to the supplementary video for additional qualitative results on individual control of the human and object modules, latent code interpolation, and composition of multiple objects.

\fi

\end{document}